\pdfoutput=1
\documentclass[11pt]{article}
\usepackage[final]{acl}

\usepackage{times}
\usepackage{latexsym}
\usepackage[T5]{fontenc}
\usepackage[utf8]{inputenc}

\usepackage{microtype}

\usepackage{inconsolata}

\usepackage{graphicx}
\usepackage{amsmath}
\usepackage{subcaption}
\usepackage[english]{babel}
\usepackage{tipa}
\usepackage{longtable}
\usepackage{booktabs}
\usepackage{array}
\usepackage{multirow}
\usepackage{amssymb}
\usepackage{adjustbox}
\usepackage{stfloats}

\usepackage{multirow}
\usepackage{xcolor}
\usepackage{makecell}
\usepackage{tabularx}
\usepackage{tabularray}
\usepackage{verbatim}
\usepackage{fancyvrb}
\title{Multi-Dialect Vietnamese: Task, Dataset, Baseline Models and Challenges}

\author{Nguyen Van Dinh\textsuperscript{1,2},
Thanh Chi Dang\textsuperscript{1,2},
Luan Thanh Nguyen\textsuperscript{1,2},
Kiet Van Nguyen\textsuperscript{1,2} \\
\textsuperscript{1}Faculty of Information Science and Engineering, University of Information Technology, \\ Ho Chi Minh City, Vietnam \\
\textsuperscript{2}Vietnam National University, Ho Chi Minh City, Vietnam \\
\texttt{\{20520657, 20520761\}@gm.uit.edu.vn} \\
\texttt{\{luannt, kietnv\}@uit.edu.vn}
}

\begin{document}
\maketitle
\begin{abstract}

Vietnamese, a low-resource language, is typically categorized into three primary dialect groups that belong to Northern, Central, and Southern Vietnam. However, each province within these regions exhibits its own distinct pronunciation variations. Despite the existence of various speech recognition datasets, none of them has provided a fine-grained classification of the 63 dialects specific to individual provinces of Vietnam. To address this gap, we introduce \textbf{V}ietnamese \textbf{M}ulti-\textbf{D}ialect (\textbf{ViMD}) dataset, a novel comprehensive dataset capturing the rich diversity of 63 provincial dialects spoken across Vietnam. Our dataset comprises 102.56 hours of audio, consisting of approximately 19,000 utterances, and the associated transcripts contain over 1.2 million words. To provide benchmarks and simultaneously demonstrate the challenges of our dataset, we fine-tune state-of-the-art pre-trained models for two downstream tasks: (1) Dialect identification and (2) Speech recognition. The empirical results suggest two implications including the influence of geographical factors on dialects, and the constraints of current approaches in speech recognition tasks involving multi-dialect speech data. Our dataset is available for research purposes\footnote{\href{https://github.com/nguyen-dv/ViMD_Dataset}{https://github.com/nguyen-dv/ViMD\_Dataset}}.

\end{abstract}

\section{Introduction}

The Vietnamese language can be divided into three major dialects (regional dialects): Northern, Central, and Southern dialect \cite{3_regiondialect_2, 3_regiondialect_1, VNese_linguistic}, each associated with distinct region and characterized by unique phonetic characteristic \citep{confusion_l_n_North, QuangNam, different_NCS_1, different_NCS_2}. However, even within these regions, the dialects unique to each province (provincial dialect) maintain noticeable differences \citep{north_central_Vietnamese, QuangNam}. Therefore, to obtain a more detailed insight into the Vietnamese dialects, we need to consider a more granular level, particularly provincial dialects, rather than limiting our examination to regional dialects alone. It is worth noting that Vietnamese is a monosyllabic language in which each word is a single syllable and words are separated by spaces. Moreover, in the Vietnamese context, the terms `accent' and `dialect' are used interchangeably \citep{different_NCS_2} and consequently, for the sake of consistency, this research adopts the term `dialect'.

In recent years, Vietnamese speech-related research has made remarkable strides; however, the issue of multi-dialectal variations within the language has posed a significant challenge \citep{ASR_challenged_by_accent, ASR_Dialect_reduce_efficiency, ASR_need_accent_dataset}. In an effort to tackle this obstacle, several multi-dialect Vietnamese corpora have been published \citep{VNSpeechCorpus, LSVSC, ViASR, VDSPEC}. However, these datasets exhibit two limitations: (1) All of the mentioned datasets encompass only three to five groups of dialects, (2) Several datasets are not publicly available.

Motivated by these two limitations, we introduce the ViMD speech dataset, a novel resource encompassing 63 provincial dialects, representing all 63 provinces of Vietnam. The audio data is curated from publicly available sources. The transcripts undergo a semi-automatic labeling process, followed by a rigorous manual verification to guarantee the quality of dataset. Furthermore, each record includes additional attributes such as speaker identification codes and gender, allowing the dataset to support various speech-related tasks.

The contributions of our study are as follows:
\begin{itemize}

    \item We release the first comprehensive multi-dialect Vietnamese speech dataset, offering a fine-grained classification of the 63 dialects, with each dialect being unique to a specific province of Vietnam. The dataset comprises 102.56 hours of audio recordings, nearly 19,000 utterances, and over 1.2 million syllables.

    \item We conduct experiments on two tasks: dialect identification (DI) and speech recognition (SR) to provide benchmarks and demonstrate the challenging nature of the dataset.

    \item Based on the experimental results, we present two in-depth analyses, including the impact of geographical factors on dialects and the limitations of the speech recognition approach when dealing with multi-dialect speech data.
    
\end{itemize}

% can be found at: github.com/gigido
\section{Basic Phonetic Structure of Vietnamese}

Vietnamese, as a monosyllabic tonal language, is structured with three key components: Initial, Final, and Tone. In particular, the Final segment is composed of three elements: Onset, Nucleus, and Coda. These components are detailed in \autoref{tab:VNese syllable}. The two mandatory elements to construct a syllable, highlighted in bold, are Tone and Nucleus. For instance, the word `bạn' (friend) exemplifies all three key components, where `b' is the Initial, `an' is the Final, and the High-Broken tone is represented by the diacritic below the Nucleus `a'. On the other hand, the word `u' (lump) has only the two mandatory elements, with `u' as the Nucleus and the Mid Tone.

There are six tones in Vietnamese: Mid Tone, High-Rising, Low-Falling, Low-Rising, High-Broken, and Low-Broken. Each tone, when combined with a syllable, conveys a distinct meaning. For instance, considering a word with the initial consonant `b' and the nucleus `a': Mid Tone (ba - three), High-Rising (bá - uncle), Low-Falling (bà - grandmother), Low-Rising (bả - poison), High-Broken (bã - waste), and Low-Broken (bạ - randomly). The differences in Pitch Contour between the tones are represented in \autoref{tab:VNese_tone} \cite{VNese_linguistic}. 

\begin{table}[h]
  \centering
  \begin{tabular}{cccc}
    \toprule
    \multicolumn{4}{c}{\textbf{Tone}} \\
    \midrule
    \multirow{2}{*}{Initial} & \multicolumn{3}{c}{Final} \\
    \cmidrule(lr){2-4}
    & Onset & \textbf{Nucleus} & Coda \\
    \bottomrule
  \end{tabular}
  \caption{Structure of Vietnamese syllable.}
  \label{tab:VNese syllable}
\end{table}


\begin{table}
\centering
\caption{Structure of Vietnamese tones.}
\label{tab:VNese_tone}
\resizebox{\linewidth}{!}{%
\begin{tblr}{
  cells = {c},
  cell{1}{1} = {r=2}{},
  cell{1}{2} = {r=2}{},
  cell{1}{3} = {c=2}{},
  hline{1,5} = {-}{0.08em},
  hline{2} = {3-4}{0.03em},
  hline{3} = {-}{0.05em},
}
\textbf{Pitch Contour} & \textbf{Flat} & \textbf{UnFlat} &             \\
                       &               & Broken          & Unbroken    \\
High                   & No mark       & High-broken     & High-rising \\
Low                    & Low-falling   & Low-rising      & Low-broken  
\end{tblr}
}
\end{table}

% \begin{table}[h]
%     \centering
%     \small
%     \begin{tabular}{@{}c{1.8cm}c{1.8cm}c{1.8cm}@{}}
%         \toprule
%         \multirow{2}{*}{\parbox{1.5cm}{\centering \textbf{Pitch} \\ \textbf{Contour}}} & \multirow{2}{*}{\textbf{Flat}} & \multicolumn{2}{c@{}}{\textbf{UnFlat}} \\
%         \cmidrule{3-4}
%         & & Broken & Unbroken \\
%         \midrule
%         High & No mark & High-broken & High-rising \\
%         Low & Low-falling & Low-rising & Low-broken \\
%         \bottomrule
%     \end{tabular}
%     \caption{Structure of Vietnamese tones.}
%     \label{tab:VNese tone}
% \end{table}

Pronunciation across provinces and regions in Vietnam has its own distinct characteristics. These differences can appear in any component of the syllable. We present more details about these variations in \autoref{sec: Appendix_A}, and their impact on the Speech Recognition task in Section \ref{sec: SR_ExperimentalResults}.
\section{Related Work}

Numerous speech corpora spanning diverse linguistic and dialectical backgrounds have been curated to facilitate dialect classification and speech recognition tasks.

\subsection{Global Corpora}
\textbf{QASR} \cite{qasr}, a large-scale Arabic speech and transcription dataset, contains 2,000 hours of 16kHz audio recordings from Aljazeera news channel. It covers 5 Arabic dialects with samples from 19,000 speakers, making it valuable for speech recognition and dialect classification research. The dataset exhibits a gender imbalance (69\% male, 6\% female), with gender unidentified for speakers having fewer than 20 audio samples.

\textbf{KeSpeech} \cite{kespeech} comprises 1,542 hours of recorded speech from 27,237 speakers in 34 cities, covering standard Mandarin and its eight subdialects. This extensive dataset supports a variety of speech processing tasks such as speech recognition, speaker recognition, and sub-dialect identification, promoting the development of multitask learning and conditional learning models. Notably, KeSpeech's parallel recording of standard Mandarin and specific sub-dialects opens new avenues for applications like dialectal voice conversion.

\textbf{STT4SG-350} \cite{stt4sg350} represents a substantial advancement in speech technology resources for Swiss German dialects. A total of 343 hours of recordings spread across seven dialects are included in this dataset, featuring 217,687 unique sentences voiced by 316 speakers. This corpus, annotated with standard German text at the sentence level, addresses the challenges of each dialect and has a large vocabulary size of 42,980.
%As the largest corpus of its kind

\textbf{Thai Dialect Corpus} \cite{thaidialect} approximately 840 hours of recordings: Thai-central with 700 hours of the main Thai dialect; Thai-dialect comprising three Thai dialects including Khummuang, Korat, and Pattani recorded from local people from three corresponding regions: North, Northeast, and South Thailand, with each dialect containing about 40 hours of data. However, this dataset faces a significant gender imbalance, with a ratio of 80\% male and 20\% female.

\textbf{Thai-Dialect} \cite{Thai_Dialect} contains speech-to-text data for 10 Thai dialects from various regions of Thailand. It includes the standard Thai-central (THA) dialect, along with northern dialects (Khummuang, Nan, Yno), northeastern dialects (Korat, Khmer, Laos), and southern dialects (Krabi, Pattani, Phangnga). All transcriptions adhere to the Thai writing standard.

\subsection{Vietnamese Corpora}

Although the Vietnamese language has numerous datasets to facilitate ASR task such as VIVOS \cite{vivos}, speech corpus by \cite{Viettel}, VinBigdata-VLSP2020-100h\footnote{\href{https://vlsp.org.vn/resources}{https://vlsp.org.vn/resources}}, FPT Open Speech Dataset (FOSD)\footnote{\href{https://data.mendeley.com/datasets/k9sxg2twv4/4}{https://data.mendeley.com/datasets/k9sxg2twv4/4}}, very few datasets incorporate regional dialectical elements.

%drive.google.com/file/d/1vUSxdORDxk-ePUt-bUVDahpoXiqKchMx

\textbf{VNSpeechCorpus} \cite{VNSpeechCorpus}, published in 2004, is considered one of the first datasets on Vietnamese dialects, divided into three region dialects of Vietnam. However, speakers representing these regions were limited to four localities: Hanoi represents the northern dialect; Nghe An, Ha Tinh represents the central dialect; and Ho Chi Minh City represents the southern dialect. The dataset consists of 100 hours of reading-style speech data, recorded in noiseless and office settings. Although the authors did not conduct dialect identification or speech recognition, they designed a vocabulary and a phonetic dictionary. The dataset is not publicly accessible.

\textbf{VDSPEC} \cite{VDSPEC}, published in 2016, comprises a duration of 45.12 hours. The authors used the dialects of Hanoi, Hue and Ho Chi Minh City to represent the Northern, Central, and Southern Vietnamese dialects, respectively. The recordings for each dialectal subset were obtained in a reading style. Furthermore, the authors performed dialect identification on this dataset using LDA and GMM models. Public access to this dataset was not granted.

\textbf{ViASR} \cite{ViASR}, released in 2023, consists of a 32-hour speech collection featuring three distinct regional dialects. The data was collected from openly accessible online sources, focusing on finance-related topics. The authors conducted baseline speech recognition experiments on transformer-based models including Whisper \cite{whisper}, Wav2vec 2.0 \cite{wav2vec2}, and MMS \cite{MMS}. This dataset has not been publicly released \footnote{Access status last verified on 21st April 2024.}.

\textbf{LSVSC} \cite{LSVSC}, published in 2024, comprises 100.5 hours of audio in a spontaneous style. The dataset includes five dialects: Northern, Central, Southern, Central Highlands, and minority ethnic group dialects. Nonetheless, the dataset shows a significant disproportion in dialects, with the Northern dialect accounting for 88.1\% of the dataset. To assess speech recognition performance on this dataset, the authors employed LAS \cite{LAS} and the Speech-Transformer Model \cite{speech_transformer} for experiments. Notably, this dataset is publicly available.% for the research community.

In summary, all the aforementioned Vietnamese corpora only categorize dialects into regional clusters. Consequently, further subdividing these regional clusters into smaller provincial dialects is an extremely challenging task. Moreover, some of these datasets are modest in size, are not publicly available, or exhibit imbalances across dialects, hindering the development of research on Vietnamese dialects. This motivates us to construct a dataset that addresses these shortcomings, spanning 102.56 hours, with relatively balanced representation across dialects in various aspects. Of paramount importance, it encompasses all 63 provincial dialects of Vietnamese, and from these 63 dialects, they can be organized into 3 regional dialects or any other dialect groupings based on the objectives of the research. \autoref{tab:dataset_related_work} presents a comparison of the multi-dialect Vietnamese speech datasets.

\begin{table*}[!t]
\centering
\small
\begin{tabular}{@{}lcrcccc@{}}
\toprule
\textbf{Dataset} & \textbf{Style} & \textbf{Duration} & \textbf{Availability Status} & \textbf{Number of Dialects} & \textbf{DI} & \textbf{SR} \\
\midrule
VNSpeechCorpus \cite{VNSpeechCorpus} & Reading & 100h & Restricted & 3 & & \\
VDSPEC \cite{VDSPEC} & Reading & 45.12h & Restricted & 3 & \checkmark &  \\
ViASR \cite{ViASR} & Spontaneous & 32h & Restricted & 3 &  & \checkmark \\
LSVSC \cite{LSVSC} & Spontaneous & 100.5h & Public & 5 &  & \checkmark \\
ViMD (our) & Spontaneous & 102.56h & Public & 3/63 & \checkmark & \checkmark \\
\bottomrule
\end{tabular}
\caption{Comparison of the multi-dialect Vietnamese speech datasets. Compared aspects include Style, Duration, Availability Status, Number of Dialects, and whether Dialect Identification (DI) or Speech Recognition (SR) was conducted.}
\label{tab:dataset_related_work}
\end{table*}
\section{ViMD Dataset}
\subsection{Data Collection}
\label{sec: Data Collection}

% 1.0
\begin{figure*}[h!]
    \centering
    \includegraphics[width=1.0\linewidth]{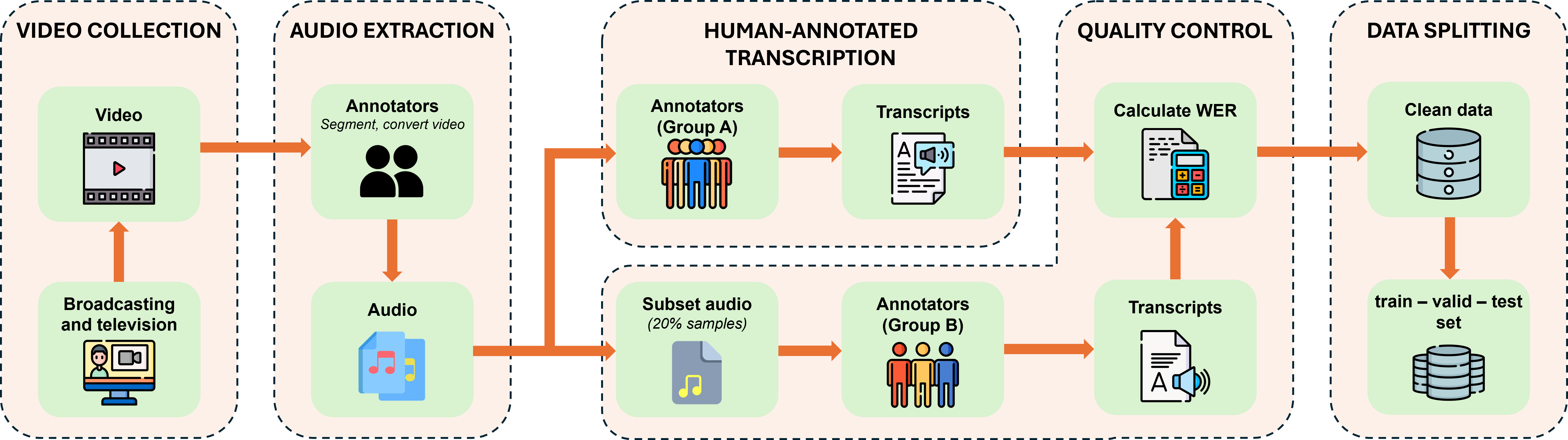}
    \caption{Data Collection Pipeline for the ViMD Dataset.}
    \label{fig:data_collection_pipeline}
\end{figure*}

We describe the process of building the ViMD according to \autoref{fig:data_collection_pipeline}. The process consists of five phases: Video collection, Audio extraction, Human-Annotated transcription, Quality control and Data splitting.

\textbf{Video Collection}. We gather videos featuring interviews with local residents from official Television and Broadcasting Station as the data for the respective provincial dialect.

\textbf{Audio Extraction}. 10 student annotators with backgrounds in information technology segment videos into short clips containing local speakers' voices with the help of the Open Source Data Labeling Platform – Label Studio \cite{LabelStudio}. Three mandatory requirements to ensure the dataset’s quality are: (1) Each segment includes only one person's voice, (2) The audio for each person is limited to 180 seconds, (3) Filtering out news presenter's audio, (4) No segment exceeds 30 seconds. Any video segments that did not meet these three requirements were eliminated. The remaining video segments are converted to audio with the .wav format, preserving their original sample rates without any standardization.

\textbf{Human-Annotated Transcription}. This phase is conducted by our annotation group A, consisting of 10 individuals with diverse linguistic backgrounds, representing the Northern, Central, and Southern regions of Vietnam. We use an semi-automated labeling procedure, starting with audio transcription generated by API of AssemblyAI \footnote{\href{https://www.assemblyai.com/}{https://www.assemblyai.com}}. Subsequent to the initial transcription, transcripts from each provincial dialect are reviewed by an individual annotator, who corrects any inaccuracies to produce refined transcripts.

\textbf{Quality Control}. To verify the quality of Group A's work, two members of Group B independently transcribe 20\% of the samples. We then use transcripts of Group B as the ground truth and calculate the Word Error Rate (WER) by comparing them to transcripts of Group A. If the WER is below 8\%, we consider the transcript valid. Otherwise, Group A must re-transcribe the entire samples of that dialect. This threshold is considered as perfect, high quality data, as outlined in \cite{Reason_8percent_WER_threshold}.

\textbf{Dataset Splitting}. We split the data into train, validation, and test sets in an 8:1:1 ratio based on duration, with the additional considerations of gender proportion and speaker exclusivity across sets.

Finally, we stored the metadata information for the audio files in JSON format, with 8 attributes described in \autoref{tab:description_key}. Further details about dataset are described in \autoref{sec: Appendix_B}. In addition, each annotator was compensated with 1.92 USD and 2.4 USD per audio hour for Audio Extraction and Human-Annotated Transcription phase, respectively.

\begin{table*}[t]
\centering
\small
\begin{tabular}{@{}llp{12cm}@{}}
\toprule
\textbf{Key} & \textbf{Data type} & \textbf{Description} \\
\midrule
set & String & The set of audio, which can take values from \{`train`, `valid`, `test`\}. \\
filename & String & Filename of the audio.\\% in the format: \begin{verbatim} {province code}_{Sequence Number of Audio}.wav \end{verbatim} \\
text & String & Transcript of the audio. \\
length & Float & Length of the audio in seconds. \\
province & String & The provincial dialect of the sample. \\
region & String & The regional dialect of the sample, which can be `North`, `Central', or `South`.\\
speakerID & String & The speaker Identification.\\% following the syntax: \begin{verbatim} spk_{province code}_{Sequence Number of Speaker} \end{verbatim} \\
gender & Int & Gender of the speaker, where 0 represents female and 1 represents male. \\
\bottomrule
\end{tabular}
\caption{Detailed Attribute Descriptions for Audio Samples in ViMD Dataset.}
\label{tab:description_key}
\end{table*}

\subsection{Dataset Statistics}

\begin{table*}[t]
\centering
\begin{tabular}{@{}lcccccccc@{}}
\toprule
& \multicolumn{4}{c}{\textbf{Per Provincial Dialect}} & \multicolumn{3}{c}{\textbf{Data Set}} & \textbf{Total}\\
\cmidrule(lr){2-5} \cmidrule(lr){6-8}
& Min. & Max. & Mean & Std. & Train & Valid. & Test & \\
\midrule
\textbf{Duration} & 89.11m & 117.98m & 97.68m & 4.18m & 81.43h & 10.26h & 10.87h & 102.56h \\
\textbf{\#record} & 263 & 363 & 301 & 21 & 15,023 & 1,900 & 2,026 & 18,949 \\
\textbf{\#speaker} & 88 & 309 & 206 & 47 & 10,291 & 1,320 & 1,344 & 12,955 \\
\textbf{\#word} & 17,038 & 24,557 & 19,669 & 1,174 & 981,391 & 125,305 & 132,471 & 1,239,167 \\
% Syllables & 17.0k & 24.5k & 19.7k & 1.2k & 981k & 125k & 132k & 1.2M \\
\textbf{\#unique-word} & 1,120 & 1,639 & 1,405 & 103 & 4,813 & 2,660 & 2,773 & 5,155 \\
\bottomrule
\end{tabular}
\caption{ViMD dataset statistics on duration, number of records, speakers, words and unique words.}
\label{tab:overall_statistics}
\end{table*}

Overall, our dataset offers a comprehensive representation of Vietnamese dialects, including 63 provincial dialects. It consists of 102.56 hours of audio recordings, with nearly 19,000 records obtained from 12,955 speakers. The accompanying transcripts consist of over 1.2 million words, with a distinct vocabulary of 5,155 unique words. The train, validation, and test sets were split in an 8:1:1 ratio based on duration, resulting in 81.43 hours, 10.26 hours, and 10.87 hours, respectively. This ratio extended to the number of records, speakers, and words as well. Notwithstanding such a ratio, the unique word counts in the validation set (2,660 unique words) and test set (2,723 unique words) does not differ significantly from from the training set (4,813 unique words), thus preserving the vocabulary diversity. Across the 63 provincial dialects, with the exception of the number of speakers exhibiting imbalance, the remaining attributes – duration, number of records, words, and unique words – are well-balanced. The relevant statistics are provided in \autoref{tab:overall_statistics}.

\begin{figure}[h]
    \centering
    \begin{minipage}{\columnwidth}
        \centering
        \includegraphics[width=0.9\columnwidth]{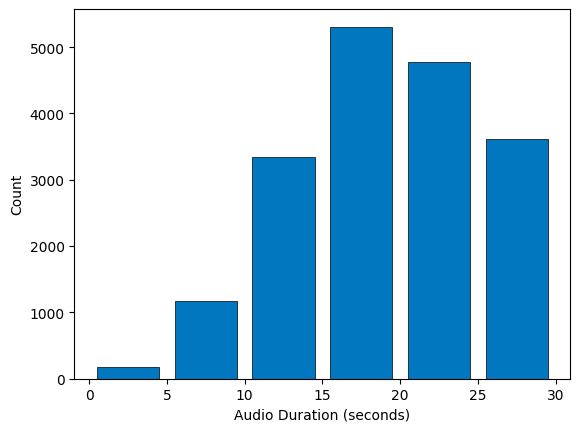}
        \caption{Distribution of Audio Duration.}
        \label{fig:Distribution of Audio Duration}
    \end{minipage}
\end{figure}

\autoref{fig:Distribution of Audio Duration} displays the distribution of audio duration, which shows a higher frequency in the range of 10 to 30 seconds. Mean and standard deviation are 19.5 and 6.2 seconds, respectively.

%0.95

\begin{figure}[t!]
    \centering
    \begin{subfigure}[b]{0.48\textwidth}
        \centering
        \includegraphics[width=\textwidth]{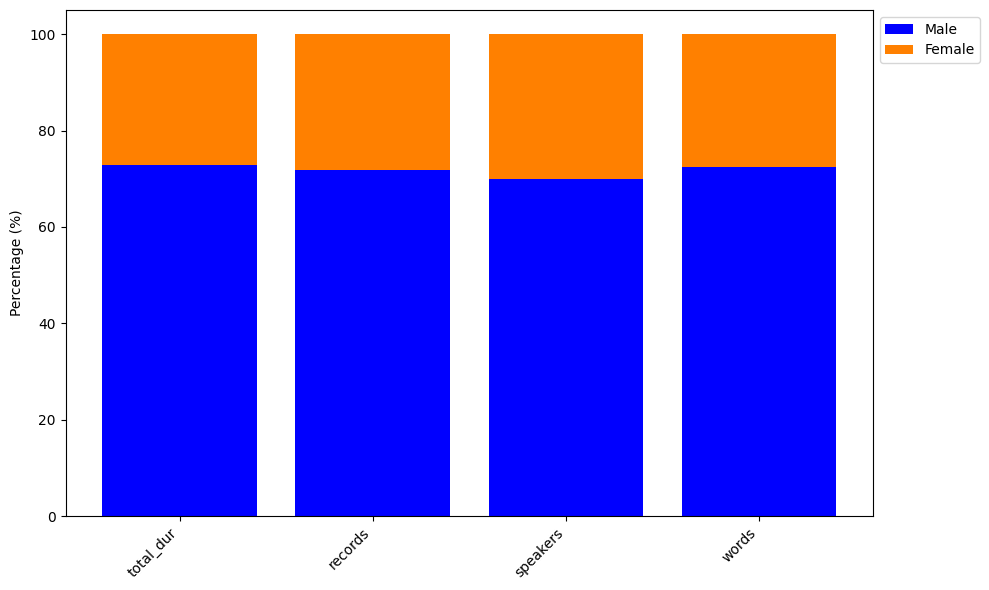}
        \caption{Gender-wise Distribution of ViMD. The blue portion at the bottom representing males and the orange portion on top representing females.}
        \label{fig:gender-dist}
    \end{subfigure}
    \hfill
    \begin{subfigure}[b]{0.48\textwidth}
        \centering
        \includegraphics[width=\textwidth]{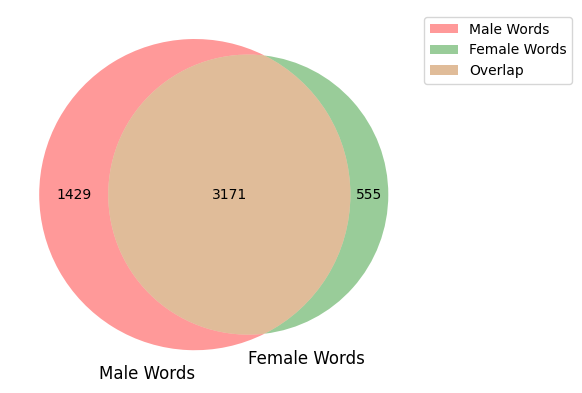}
        \caption{Gender unique word counts and overlap.}
        \label{fig:venn-diagram}
    \end{subfigure}
    \caption{(a) Gender-wise Distribution of ViMD, and (b) Gender unique word counts and overlap.}
    \label{fig:combined}
\end{figure}

\autoref{fig:gender-dist} reveals that in our dataset, the duration, records, speakers, and syllables for males are three times greater than those for females. Meanwhile, \autoref{fig:venn-diagram} illustrates a considerable overlap in unique words between males and females, with 3,171 overlaps out of 4,600 unique male words and 3,726 unique female words. These findings suggest that while there are significant differences in duration and other attributes between males and females, there is a notable diversity of word diversity between the two genders.

\section{Experiments and Results}

We conduct experiments on our dataset through two tasks encompassing  \textbf{Dialect identification} and \textbf{Speech recognition}. The process is presented through the following sections: experimental design, baselines evaluation metrics, data pre-processing, and results.

%                               ----- EXPERIMENTAL DESIGN -----
\subsection{Experimental Design}

\textbf{Dialect Identification.} We use only audio data to fine-tune pre-trained models, proceeding through five experiments.

\begin{itemize}
    \item In the initial task, we categorize dialects into three labels including Northern, Central, and Southern, based on the `region' attribute specified in Table 2. This experiment, employing the entire 102.56 hours of data, is referred to as \textbf{[DI\_VN\_3]}.

    \item We divide our dataset into three sub-datasets, comprising 25 provinces in the Northern region (40.59 hours), 19 provinces in the Central region (31.47 hours), and 19 provinces in the Southern region (30.5 hours). Subsequently, we perform provincial dialect classification on these three sub-datasets, called \textbf{[DI\_North]}, \textbf{[DI\_Central]}, and \textbf{[DI\_South]}, respectively.

    \item The last experiment in our DI task is denoted as \textbf{[DI\_VN\_63]}, in which we undertake the comprehensive classification of 63 labels corresponding to the 63 provincial dialects, utilizing the full 102.56 hours of data.
    
\end{itemize}

\textbf{Speech Recognition.} For the SR task, we utilize both audio data and accompanying transcripts. We carry out four experiments, with two stages in each experiment including direct inference from the models, and inference after fine-tuning on our data.

\begin{itemize}

    \item Similar to the DI task, we employ three sub-datasets representing the Northern, Central, and Southern region of Vietnam. The experiments are respectively named \textbf{[SR\_North]}, \textbf{[SR\_Central]}, and \textbf{[SR\_South]}.

    \item For the most comprehensive experiment of SR task,  we leverage the entirety of the dataset to perform speech recognition across all dialects of Vietnam. This task is named \textbf{[SR\_VN\_63]}.

\end{itemize}

%                               ----- BASELINES -----
\subsection{Baseline Models}

To evaluate this challenging dataset, we conduct experiments on both tasks using the state-of-the-art transformer-based pre-trained models. These models achieve impressive results in speech-related tasks, particularly in the field of automatic speech recognition.

\textbf{Wav2vec 2.0} \cite{wav2vec2} is a state-of-the-art self-supervised Learning model. It uses CNN, transformer, and quantization modules. During training, raw audio is mapped to quantized speech representations used in a contrastive task. Cosequently, the model is fine-tuned on labeled data for transcription. We use wav2vec2-base-vi\footnote{\href{https://huggingface.co/nguyenvulebinh/wav2vec2-base-vi}{huggingface.co/nguyenvulebinh/wav2vec2-base-vi}}, wav2vec2-large-vi\footnote{\href{https://huggingface.co/nguyenvulebinh/wav2vec2-large-vi}{huggingface.co/nguyenvulebinh/wav2vec2-large-vi}} for DI task and wav2vec2-base-vietnamese\footnote{\href{https://huggingface.co/dragonSwing/wav2vec2-base-vietnamese}{huggingface.co/dragonSwing/wav2vec2-base-vietnamese}}, wav2vec2-base-vietnamese-160h\footnote{\href{https://huggingface.co/khanhld/wav2vec2-base-vietnamese-160h}{huggingface.co/khanhld/wav2vec2-base-vietnamese-160h}}, wav2vec2-base-vietnamese-250h\footnote{\href{https://huggingface.co/nguyenvulebinh/wav2vec2-base-vietnamese-250h}{huggingface.co/nguyenvulebinh/wav2vec2-base-vietnamese-250h}}, wav2vec2-base-vi-vlsp2020\footnote{\href{https://huggingface.co/nguyenvulebinh/wav2vec2-base-vi-vlsp2020}{huggingface.co/nguyenvulebinh/wav2vec2-base-vi-vlsp2020}} for SR task.

\textbf{XLSR} \cite{xlsr} and \textbf{XLS-R} \cite{xls-r} are multilingual extensions of wav2vec 2.0 for cross-lingual speech recognition. XLSR uses a shared quantizer for pretraining on diverse languages. XLS-R is an enhanced version with larger models and more language coverage. We apply XLSR and XLS-R for only DI task, including wav2vec2-xls-r-300m\footnote{\href{https://huggingface.co/facebook/wav2vec2-xls-r-300m}{huggingface.co/facebook/wav2vec2-xls-r-300m}} and wav2vec2-large-xlsr-53\footnote{\href{https://huggingface.co/facebook/wav2vec2-large-xlsr-53}{huggingface.co/facebook/wav2vec2-large-xlsr-53}}.

\textbf{Whisper} \cite{whisper} is OpenAI's advanced multilingual ASR system using self-supervised learning on 680,000 hours of speech data. It employs transformer models with attention, utilizing multi-task learning. The encoder processes audio spectrograms, and the decoder generates transcripts from the audio features. PhoWhisper \cite{PhoWhisper} is a fine-tuned version of Whisper, trained on an 844-hour Vietnamese speech dataset. We employ whisper\text{\footnotesize\raisebox{-0.5ex}{base}} and PhoWhisper\text{\footnotesize\raisebox{-0.5ex}{base}} for both DI and SR tasks.

We provide detailed information about the computational resources and hyperparameter settings in \autoref{sec: Appendix_C}.

%                               ----- EVALUATION METRICS -----

\subsection{Evaluation Metrics}

F1-macro \cite{f1_macro} is utilized for the tasks of dialect identification, while Word Error Rate (WER) \cite{WER} is employed for speech recognition tasks. Detailed information regarding these two metrics will be provided in \autoref{sec: Detail_Metrics}.

%                               ----- DATA PRE-PROCESSING -----

\subsection{Data Pre-processing}

All audio files are resampled to a 16kHz sampling rate and converted to mono channel. For the dialect identification task, files with a duration under 10 seconds are kept intact, whereas longer files are segmented into segments not exceeding 10 seconds. This segmentation approach stems from established practices in related research \cite{Reason_DI_Split10s_cite1, Reason_DI_Split10s_cite2}. As for the speech recognition task, the text data undergoes several common preprocessing techniques, such as converting to lowercase and removing punctuation marks.

%                              ----- DIALECT IDENTIFICATION RESULTS -----

\begin{table*}[!t]
\centering
\small
\begin{tabular}{lcccccc}
\toprule
& \textbf{\#Params} & \textbf{Vietnam} & \textbf{North} & \textbf{Central} & \textbf{South} & \textbf{Vietnam} \\
& & [DI\_VN\_3] & [DI\_North] & [DI\_Central] & [DI\_South] & [DI\_VN\_63] \\
\midrule
\textbf{Model/Num. of labels} & & \textbf{3} & \textbf{25} & \textbf{19} & \textbf{19} & \textbf{63} \\
\midrule
\texttt{wav2vec2-base-vi} & 95M & 0.9102 & 0.4322 & 0.5863 & \textbf{0.3560} & 0.3522 \\
\texttt{wav2vec2-large-vi} & 317M & \textbf{0.9147} & 0.4229 & 0.5981 & 0.3528 & 0.3570 \\
\texttt{wav2vec2-xls-r-300m} & 300M & 0.8901 & 0.3216 & 0.3282 & 0.2787 & 0.3728 \\
\texttt{wav2vec2-large-xlsr-53} & 300M & 0.8736 & 0.2830 & 0.1990 & 0.2440 & 0.3255 \\
\texttt{whisper\text{\footnotesize\raisebox{-0.5ex}{base}}} & 74M & 0.8559 & 0.4336 & 0.5854 & 0.3383 & 0.3976 \\
\texttt{PhoWhisper\text{\footnotesize\raisebox{-0.5ex}{base}}} & 74M & 0.8697 & \textbf{0.4470} & \textbf{0.6251} & 0.3257 & \textbf{0.4107} \\
\bottomrule
\end{tabular}
\caption{Dialect identification experimental results with F1-macro metric.}
\label{tab:DI_result}
\end{table*}

\begin{table*}[!t]
\centering
\small
\begin{tabular}{lccccc}
\toprule
\textbf{Model} & \textbf{\#Params} & \textbf{North} & \textbf{Central} & \textbf{South} & \textbf{Vietnam} \\
& & [SR\_North] & [SR\_Central] & [SR\_South] & [SR\_VN\_63] \\
\midrule
\multicolumn{6}{c}{\textbf{w/o Fine-tuned}} \\
\midrule
\texttt{wav2vec2-base-vietnamese} & 95M & 0.2032 & 0.2728 & 0.2248 & 0.2307 \\
\texttt{wav2vec2-base-vietnamese-160h} & 95M & 0.2750 & 0.3812 & 0.3093 & 0.3174 \\
\texttt{wav2vec2-base-vietnamese-250h} & 95M & 0.1498 & 0.2097 & 0.1724 & 0.1747 \\
\texttt{wav2vec2-base-vi-vlsp2020} & 95M & \textbf{0.1364} & \textbf{0.1926} & \textbf{0.1481} & \textbf{0.1568} \\
% \texttt{wav2vec2-large-vi-vlsp2020} & \textbf{0.1348} & \textbf{0.1865} & \textbf{0.1642} & \textbf{0.1603} \\
\texttt{whisper\text{\footnotesize\raisebox{-0.5ex}{base}}} & 74M & 0.2637 & 0.3991 & 0.2946 & 0.3138 \\
\texttt{PhoWhisper\text{\footnotesize\raisebox{-0.5ex}{base}}} & 74M & 0.1496 & 0.2415 & 0.1787 & 0.1861 \\
\midrule
\multicolumn{6}{c}{\textbf{Fine-tuned}} \\
\midrule
\texttt{wav2vec2-base-vietnamese} & 95M & 0.1464 & 0.2027 & 0.1772 & 0.1580 \\
\texttt{wav2vec2-base-vietnamese-160h} & 95M & 0.1670 & 0.2456 & 0.2051 & 0.1749 \\
\texttt{wav2vec2-base-vietnamese-250h} & 95M & 0.1229 & \textbf{0.1715} & 0.1526 & 0.1356 \\
\texttt{wav2vec2-base-vi-vlsp2020} & 95M & \textbf{0.1217} & 0.1719 & 0.1508 & \textbf{0.1224} \\
% \texttt{wav2vec2-large-vi-vlsp2020} & \textbf{0.1232} & \textbf{0.1681} & 0.1487 & \textbf{0.1277} \\
\texttt{whisper\text{\footnotesize\raisebox{-0.5ex}{base}}} & 74M & 0.2005 & 0.2789 & 0.2089 & 0.1993 \\
\texttt{PhoWhisper\text{\footnotesize\raisebox{-0.5ex}{base}}} & 74M & 0.1320 & 0.1826 & \textbf{0.1354} & 0.1630 \\
\bottomrule
\end{tabular}
\caption{Speech recognition experimental results with WER metric.}
\label{tab:asr-result}
\end{table*}

\subsection{Dialect Identification Experimental Results}

The results of the DI experiment are shown in \autoref{tab:DI_result}. Our analysis focuses on two aspects: first, we compare the performance across the different models used. Second, we look at how the models perform when tested on experiments with different audio data and label categories.

In the [DI\_VN\_3], the wav2vec 2.0 family achieves the highest F1-macro scores, with 91.02\% for the base model and 91.47\% for the large model. However, in the most challenging task - [DI\_VN\_63], their performance is relatively poor, with scores of only 35.22\% for the base model and 35.28\% for the large model. Interestingly, the whisper model group exhibits the opposite trend, with the lowest F1-macro scores in the [DI\_VN\_3] (85.59\% for whisper-base and 86.97\% for phowhisper-base) but the highest scores in the [DI\_VN\_63] (39.76\% for whisper-base and 41.07\% for phowhisper-base). The XLSR and XLS-R models' F1-macro scores are average across both tasks, with the exception of wav2vec2-large-xlsr-53, which has the lowest score of 32.55\% in the [DI\_VN\_63]. Regarding categorizing provincial dialects within each regional dialect, the wav2vec and whisper groups outperform the XLSR and XLS-R groups. PhoWhisper performs the best with 44.70\% for the [DI\_North] and 62.51\% for the [DI\_Central], while wav2vec2-base-vi achieves the best 35.60\% for the [DI\_South].

Out of the five experiments we conduct, the [DI\_VN\_3] is the simplest with the fewest labels, and the models achieve very high F1-macro scores, all above 85\%. For the identification of provincial dialects within each regional dialect, the provinces in the Central region [DI\_Central] seem to have the most distinct characteristics, with the highest F1-macro score of 62.31\%, followed by the North [DI\_North] at 44.70\%, and the model performs the worst for the Southern region [DI\_South] with only 35.60\%. When using all 63 provincial dialects for the [DI\_VN\_63], the F1 score is 41.07\%.

These experiments established a benchmark for this dataset. Simultaneously, the results also demonstrate the challenging nature of the DI task on this dataset.

%                              ----- SPEECH RECOGNITION RESULTS -----

\subsection{Speech Recognition Experimental Results}
\label{sec: SR_ExperimentalResults}

\textbf{Model Performance.} The outcomes of the SR task are summarized in \autoref{tab:asr-result}. We analyze the discrepancy between before and after fine-tuning, based on two main criteria: performance across different models and different experiments.

All direct inference results yield lower performance compared to fine-tuning, except for the case of wav2vec2-base-vi-vlsp2020 in the [SR\_South] experiment where fine-tuning leads to a 0.27\% higher WER. The model that shows the best improvement across all tasks after fine-tuning is wav2vec2-base-vietnamese-160h, although this improvement is still not sufficient to outperform other models. Prior to fine-tuning, wav2vec2-base-vi-vlsp2020 demonstrates superiority by achieving the best results across all experiments. However, after fine-tuning, wav2vec2-base-vietnamese-250h performs comparably, sometimes outperforming and sometimes underperforming wav2vec2-base-vi-vlsp2020, with the highest gap being only 1.32\% across all experiments. phowhisper-base shows significant improvement in [SR\_South], outperforming other models and achieving a result of 13.54\% in this experiment.

When fine-tuning on the entire dataset in the [SR\_VN\_63] experiment, most models show the best improvement, which can be influenced by the large data quantity. The best result on [SR\_VN\_63] is 12.24\% with the wav2vec2-base-vi-vlsp2020 model. When comparing the three experiments with smaller data sizes, [SR\_North], [SR\_Central], and [SR\_South], both before and after fine-tuning, [SR\_Central] always has the highest WER while [SR\_North] has the lowest WER across all experiments. However, an encouraging observation is that [SR\_Central] shows the most significant improvement, while [SR\_North] exhibits the least improvement after fine-tuning. This suggests that although the results for [SR\_Central] are not yet optimal, our dataset has a considerably positive impact on the current models in the Central region of Vietnam. The best results obtained after fine-tuning for [SR\_North], [SR\_Central], and [SR\_South] are 12.17\%, 17.15\%, and 13.54\%, respectively.

In general, our dataset helped improve the performance of models in the SR task. Concurrently, it has also poses challenges for models in dealing with certain specific dialects.

\textbf{The Effect of Dialectal Variations on SR Performance.} We select the best-performing model based on WER to analyze the results from four dialectal ASR experiments. We assert that using dialect-specific datasets significantly reduces spelling errors. Vietnam's linguistic diversity, with each province having at least one unique dialect, is a key factor in recognition errors. Our analysis reveals that provinces in the Northern, Central, and Southern regions often display similar dialect-specific spelling mistakes. These errors are compiled in \autoref{tab: Appendix_SR_Error_Vocab}, along with model predictions before and after fine-tuning on our dataset.

The Northern region of Vietnam typically has the clearest pronunciation. However, some local pronunciation mistakes are still prevalent, particularly the confusion between the syllables `n' and `l', as seen in `linh bình' for `ninh bình' and `việc nàm' for `việc làm'. Model performance significantly improved after fine-tuning, as evidenced by a notable reduction in the WER metric.

In the Central region, the accent exhibits the most distinct variations, resulting in a significantly higher WER in our experiments compared to other regions. Certain vowels undergo notable changes; for example, `a' becomes `o' or `a' morphs into `e', turning `thắt chặt' into `thất chẹt', `lắng nghe' into `lấn nghe', `bán' into `bón', and `năm' into `nem'. In provinces like Hue (74) and Quang Tri (75), question tones tend to be pronounced with a heavier inflection, such as `phát triện' instead of `phát triển'.

In the Southern region, words beginning with the letters `v' and `d' are commonly mispronounced as beginning with `d', for example, `dội dả' instead of `vội vã'; and `tr' is often misheard as `ch', like `nổi chội' instead of `nổi trội'. Additionally, final consonants `n' and `ng' are frequently confused as `ng', `c' and `t' are both misinterpreted as `c', such as `bạng' for `bạn', `các' for `cát'. There are also other errors such as unclear pronunciation of complex words or failure to distinguish between different but relatively similar tonal marks.
\section{Dicussion}
\label{sec: Dicussion}

Our analysis of the results yields two notable findings encompassing (1) Geographical influences on dialects and (2) Multi-Dialect data challenges for speech recognition approach. The following are our conjectures based on experimental results. The cause could also stem from the training data for pre-trained models or other factors.

\subsection{Geographical Influences on Dialects}

The detailed confusion matrix for the DI task and the table of province codes are presented in the \autoref{sec: Appendix_D}; a map of Vietnam is also included\footnote{\href{https://bandovn.vn/vi/page/mau-ban-do-hanh-chinh-nuoc-cong-hoa-xa-hoi-chu-nghia-viet-nam-181}{https://bandovn.vn/vi/page/mau-ban-do-hanh-chinh-nuoc-cong-hoa-xa-hoi-chu-nghia-viet-nam-181}}. In the [DI\_VN\_63] experiment, 12 provinces achieved F1-macro scores of 0.6 or above. What is truly remarkable is that 10 (coded as 17 - Thai Binh, 18 - Nam Dinh, 35 - Ninh Binh, 37 - Nghe An, 73 - Quang Binh, 76 - Quang Ngai, 77 - Binh Dinh, 78 - Phu Yen, 86 - Binh Thuan, 59 - Ho Chi Minh) out of these 12 provinces are coastal regions, while the remaining 2 provinces (coded as 28 - Hoa Binh and 98 - Bac Giang) are only one province away from the sea. This observation underscores the potential influence of coastal factors on the unique features of local speech patterns.

A noteworthy finding in the Central region is that although the highest DI scores ([DI\_Central]) demonstrate the highly distinctive nature of provincial dialects within the region, the SR result ([SR\_Central]) for the Central provinces are the poorest. The Central region's narrow and latitudinally elongated shape, unlike the Northern and Southern regions, could be a potential cause for this characteristic.

\subsection{Multi-Dialect data challenges for speech recognition approach}

We select the wav2vec2-base-vi-vlsp2020 model as it exhibits the best performance on the [SR\_VN\_63] experiment. We calculate the WER for each regional dialect in [SR\_VN\_63] and compare it with the corresponding WER on [SR\_North], [SR\_Central], and [SR\_South]. The improvements over training on a entire dataset are 1.86\%, 3.07\%, and 2.34\% for the Northern, Central, and Southern dialects, respectively. Details on the WER differences for each provincial dialect are listed in \autoref{sec: Appendix_D}. These differences demonstrate that despite training the model on a combined dataset containing various dialects with a duration approximately 2-3 times larger than the individual datasets, the performance gain over training on separate dialects is relatively small. The findings indicate a need for more effective methods of cross-dialect knowledge transfer, rather than merely aggregating the dialects and training on the combined dataset as a separate dataset.
\section{Conclusion}

We introduce ViMD, a novel dataset covering all 63 provincial dialects in Vietnam. We carry out experiments on the two tasks of dialect identification and speech recognition, employing various state-of-the-art models to establish baseline benchmarks. The results facilitate a more in-depth investigation of dialects, including the impact of geographical factors on dialectal variations, and pose challenges for speech recognition models in tackling the multi-dialect aspect of the Vietnamese language. In addition, we hope that our process will help expand both the scale and quality of the other datasets. We also expect that this dataset will facilitate future research aimed at enhancing the performance of DI and SR tasks, as well as other related speech tasks, especially for the Vietnamese language.

\section*{Limitations}

Although the majority of residents within a given province tend to speak the local dialect, a minority who have previously resided in other regions retain their original dialectal forms, even giving rise to "hybrid dialects". The audio duration of the dataset is quite modest at only 102.56 hours. The transcripts contained within the dataset show a few inaccuracies resulting from regional pronunciation patterns as well as vocabulary highly specific to certain areas. Additionally, there is a gender disparity, with the number of male speakers being three times that of female speakers. While we have not yet categorized the audio content by topic, the majority of the recordings are from television news programs, which typically address a broad spectrum of issues from daily life.

In our experiments, we primarily employed base version of pre-trained models, employing only a few large version of pre-trained models owing to constraints in computational resources, thus limiting our ability to conduct a comprehensive assessment of state-of-the-art model capabilities. Subsequent investigations will potentially include the use of larger versions, notably Wav2vec2-BERT \cite{Wav2Vec2_BERT} and MMS-1B \cite{MMS}.
\section*{Ethics Statement}

The entirety of the data utilized in this research study is from publicly available sources, ensuring no infringement of privacy rights. All data has been published by the 63 Television and Broadcasting Stations corresponding to all 63 provinces of Vietnam, guaranteeing the verification of all included content. Our study is aimed at furthering research efforts into the regional dialects found throughout Vietnam, and it is not intended to target any specific individuals or organizations.

\section*{Acknowledgement}
This research was supported by The VNU-HCM University of Information Technology’s Scientific Research Support Fund.

\bibliography{acl_latex}

\clearpage

\appendix
\label{sec:appendix}
% \section{Vietnamese Language}
\section{Linguistic Variations Across Dialects}
\label{sec: Appendix_A}

The diversity across the three regions of Vietnam is reflected in the pronunciation of syllable elements \cite{VNese_linguistic}:
\begin{itemize}
    \item Initial consonant: The Northern dialect possesses the smallest number of initial consonants at 20, followed by the Southern dialect with 21, and the Central dialect has the highest count with 23 initial consonants.

    \item Tone: the Northern dialect has the most tones, with 6, while the Central and Southern dialects have 5 tones each.

    \item Final consonant: The Northern dialect features 10 final consonants, the Central dialect also has 10 final consonants, and the Southern dialect has 8.

    \item Vowel: The pronunciation of vowels adheres to specific word contexts.
    
\end{itemize}

Differences in pronunciation can lead to distortions in meaning. In the Northern region, some areas interchange the pronunciation of initial consonant `l' and `n' \cite{confusion_l_n_North}. For instance, the word `lầm' (mistake) might be pronounced as `nầm' (breast of a mammal). Quang Nam province, representing the Central region, showcases a phonetic shift where the vowel `a' is pronounced as `o' \cite{QuangNam}, seen in the pronunciation of `tám' (eight) as `tóm' (catch). In the Southern region, there's a tendency to pronounce the final consonants `n' and `ng' similarly \cite{LSVSC}, as evidenced by the identical pronunciation of `lươn' (eel) and `lương' (salary). The differences in pronunciation are not only at the regional level, but they also exist among different provinces within the same region. For example, within the Central dialect, provinces in the North-Central region pronounce the letter `gi' as \textipa{[z]}, whereas some provinces in the South-Central region (represented by Quang Nam) pronounce it as \textipa{[j]}. Furthermore, within a province, there are intra-provincial differences in pronunciation, illustrated by variations in the pronunciation of `bật lửa' (lighter) across districts in Nghe An province \cite{north_central_Vietnamese}.

Besides a word having multiple pronunciations, different regions also have distinct words expressing the same meaning. \autoref{tab: Appendix_SameMeaning_DifferentWords} illustrates some of the varying words across regions \cite{different_NCS_1, different_NCS_2}.

\begin{table}[t]
\centering
\begin{adjustbox}{max width=\textwidth}
\begin{tabular}{@{}c|c|c|c@{}}
\toprule
\textbf{\fontsize{10.5}{20}\selectfont Northern} & \textbf{\fontsize{10.5}{20}\selectfont Central} & \textbf{\fontsize{10.5}{20}\selectfont Southern} & \textbf{\fontsize{10.5}{20}\selectfont Meaning}\\
\midrule
\fontsize{10.5}{20}\selectfont bố, thầy & \fontsize{10.5}{20}\selectfont bọ & \fontsize{10.5}{20}\selectfont ba, tía & \fontsize{10.5}{20}\selectfont father \\
\fontsize{10.5}{20}\selectfont u, mẹ & \fontsize{10.5}{20}\selectfont mế, bầm, mạ & \fontsize{10.5}{20}\selectfont má, me & \fontsize{10.5}{20}\selectfont mother \\
\fontsize{10.5}{20}\selectfont chúng tui & \fontsize{10.5}{20}\selectfont bầy tui & \fontsize{10.5}{20}\selectfont tụi tui & \fontsize{10.5}{20}\selectfont we \\
\fontsize{10.5}{20}\selectfont mày & \fontsize{10.5}{20}\selectfont mi & \fontsize{10.5}{20}\selectfont mầy & \fontsize{10.5}{20}\selectfont you \\
\fontsize{10.5}{20}\selectfont gì & \fontsize{10.5}{20}\selectfont chi & \fontsize{10.5}{20}\selectfont gì & \fontsize{10.5}{20}\selectfont what \\
\fontsize{10.5}{20}\selectfont đâu thế & \fontsize{10.5}{20}\selectfont mô rứa & \fontsize{10.5}{20}\selectfont đâu vậy & \fontsize{10.5}{20}\selectfont where \\
\fontsize{10.5}{20}\selectfont thế nào & \fontsize{10.5}{20}\selectfont răng & \fontsize{10.5}{20}\selectfont sao & \fontsize{10.5}{20}\selectfont how \\
\bottomrule
\end{tabular}
\end{adjustbox}
\caption{Variations in Vietnamese Words with the Same Semantics Across Regions.}
\label{tab: Appendix_SameMeaning_DifferentWords}
\end{table}
\section{Dataset}
\label{sec: Appendix_B}
\subsection{Data Construction}

In the audio transcription task described in Section \ref{sec: Data Collection}, we followed specific guidelines to ensure our transcripts were clear and consistent: (1) Numerals were converted to their word form, (2) Units of measurement were phonetically transcribed into Vietnamese, and (3) Local vernacular terms were preserved without modification. The distinction between a sentence written in the common language and our prescribed transcript format will be illustrated in \autoref{tab: Sample_format_transcript}. In the table, \textbf{\textcolor{red}{red}} highlights Guideline (1), \textbf{\textcolor{blue}{blue}} for Guideline (2), and \textbf{\textcolor{green}{green}} for Guideline (3).

Although there is a clear process for transcribing and quality control, some errors still exist in the dataset. Common errors in transcripts typically stem from the use of local vocabulary, which is often spoken rather than written, leading to inaccuracies. Additionally, mistakes frequently occur with proper names of villages, towns, districts, or individuals due to annotators' unfamiliarity with these locations. To address these issues, we propose researching and providing a list of villages, towns, cities and districts for annotators. Furthermore, referring to online sources or Vietnamese dictionaries can help ensure accurate transcription of dialectal terms, allowing for better alignment between spoken and written language. However, the error rate is limited to under 8\% word error rate.

\begin{table}[h]
    \centering
    \begin{tabular}{p{0.23\linewidth}|p{0.60\linewidth}}
        \toprule
        \makecell[l]{\textbf{Commonly} \\ \textbf{written}} & 
        \makecell[l]{
            Chúng tôi đã \textbf{\textcolor{green}{làm}} được trên \textbf{\textcolor{red}{50}} \\ \textbf{\textcolor{blue}{ha}} lúa.
        } \\
        \midrule
        \makecell[l]{\textbf{Transcript}} & 
        \makecell[l]{
            Chúng tôi đã \textbf{\textcolor{green}{mần}} được trên \\ \textbf{\textcolor{red}{năm chục}} \textbf{\textcolor{blue}{héc ta}} lúa.
        } \\
        \midrule
        \makecell[l]{\textbf{English}} & 
        \makecell[l]{
            We have \textbf{\textcolor{green}{cultivated}} over \textbf{\textcolor{red}{50}} \\ \textbf{\textcolor{blue}{hectares}} of rice paddies.
        } \\
        \bottomrule
    \end{tabular}
    \caption{Divergences Between Common Writing and Transcript Format.}
    \label{tab: Sample_format_transcript}
\end{table}

\autoref{tab:description_key} presents the 8 attributes associated with each audio sample in our dataset. Below is an example of a sample. The filename follows the syntax \{province code\}\_\{Sequence Number of Audio\}, and similarly, the speaker identification code adheres to the syntax spk\_\{province code\}\_\{Sequence Number of Speaker\}.

\small % Set the font size to small
\begin{verbatim}
{
    "set": "train",
    "filename": "19_0001.wav",
    "text": "Vật dụng để phòng chống cháy nổ ở khu 
    vực này vẫn còn thiếu rất là nhiều.",
    "speakerID": "spk_19_0001",
    "gender": 1,
    "length": 5.244
}
\end{verbatim}
\normalsize % Reset to normal size after the verbatim block

\autoref{tab: Statistic_province} presents a comprehensive list of all 63 provinces in Vietnam, including the province name in Vietnamese, the region to which the province belongs, the province code and several other attributes. The province codes are assigned based on the vehicle registration plate designations for automobiles and motorcycles in Vietnam, as regulated by the Vietnamese Government\footnote{\href{https://congbao.chinhphu.vn/noi-dung-van-ban-so-58-2020-tt-bca-31631}{https://congbao.chinhphu.vn/noi-dung-van-ban-so-58-2020-tt-bca-31631}}. For provinces with multiple codes, we have selected a representative code.

% Cổng Thông tin điện tử Chính phủ
% www.chinhphu.vn
% Thông tư 58/2020/TT-BCA
% https://congbao.chinhphu.vn/noi-dung-van-ban-so-58-2020-tt-bca-31631?cbid=31758

\subsection{Dataset Additional Statistics}
\label{sec: Appendix_Dataset_Statistics}

\begin{figure}[h]
    \centering
    \rotatebox{90}{\includegraphics[width=0.65\textheight]{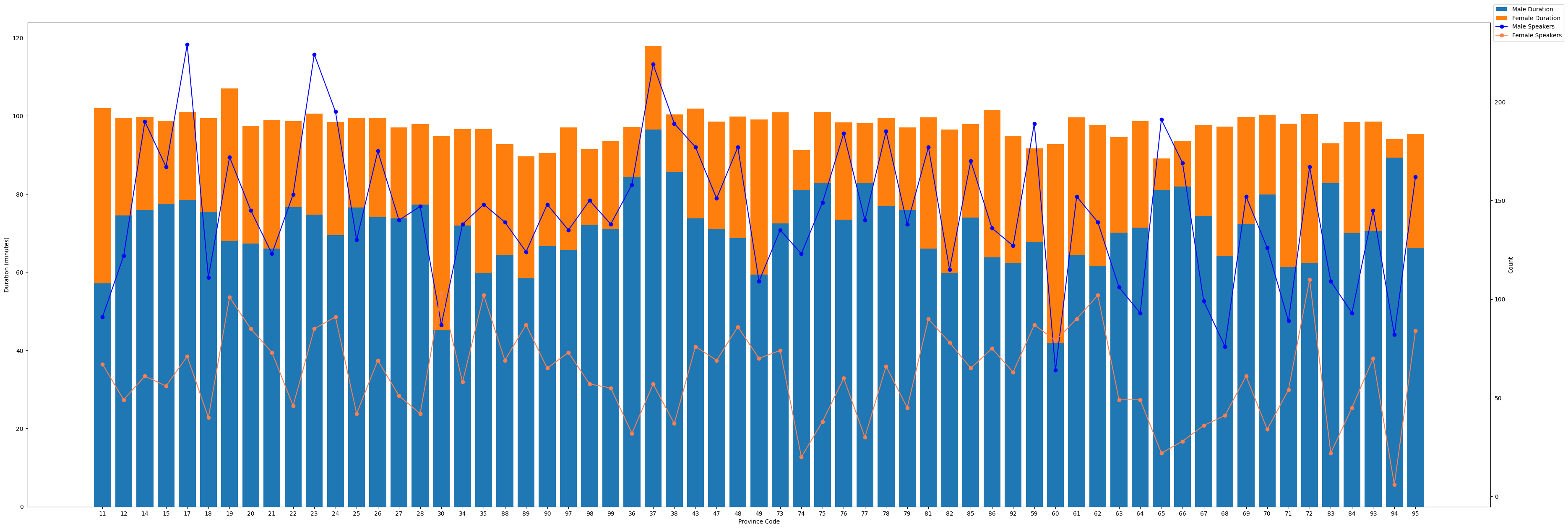}}
    \caption{Comparison of Duration and Number of Speakers Between Genders.}
    \label{fig:Duration_Spekaer_2gender}
\end{figure}

\textbf{Statistics by Provincial Dialects and Gender}. \autoref{fig:Duration_Spekaer_2gender} presents a stacked visualization of the duration across 63 provincial dialects in Vietnam, with the blue area at the bottom reflecting the duration for males, and the area above representing the duration for females. Accompanying these bars are two lines: one with blue markers depicting the number of male speakers, and another with orange markers indicating the female speaker count. While the total duration appears relatively uniform across the provincial dialects, the figure highlights a significant disparity in duration and speaker count between the two genders. \autoref{tab: Statistic_province} illustrates the number of words and the number of unique words across 63 provincial dialects, with the blue line representing the word count and the orange line indicating the count of unique words.

\begin{figure}[b]
    \centering
    \includegraphics[width=0.48\textwidth]{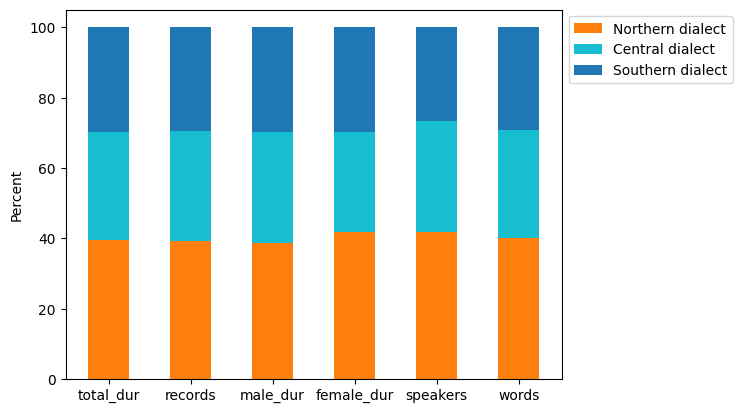}
    \caption{Comparison of Duration and Number of Speakers Between Genders.}
    \label{fig:Appendix_3region_Statistic}
\end{figure}

\begin{figure}[h]
    \centering
    \includegraphics[width=0.48\textwidth]{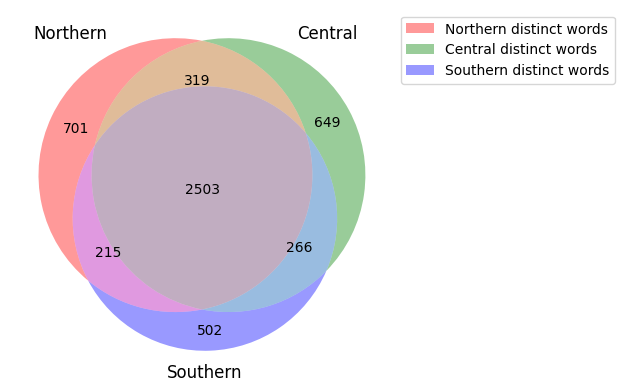}
    \caption{Words and Unique Words Count Across Provinces.}
    \label{fig:Appendix_3region_VennDiagram}
\end{figure}

\textbf{Statistics by Regional Dialects}. The statistical representation in \autoref{fig:Appendix_3region_Statistic} highlights the comparison among the three regions concerning total duration (total\_dur), number of records (records), duration of male speakers (male\_dur), duration of female speakers (female\_dur), number of speakers (speakers), and number of words (words). The distribution among the three regions appears to be relatively balanced, with only a slight predominance in the North, attributed to the larger number of provinces (25) compared to the Central (19) and Southern (19) regions. \autoref{fig:Appendix_3region_VennDiagram} shows the vocabulary overlap among regions. The intersection among all three regions is small, with only 2506 words out of 5167. Adjacent regions like Northern and Central, or Central and Southern, have more overlap than the distant Northern and Southern regions. The Northern and Central dialects have a quite similar number of unique words, 697 and 662 respectively, while the Southern dialect has fewer unique words, with 504.

\textbf{Lexical Statistics by Gender} We have listed the 6 most frequently used words for each gender, excluding one word that was a proper name. Here they are: For males: (1) ‘cồn’ (nồng độ cồn - Alcohol Concentration), (2) ‘loạt’ (đồng loạt – simultaneously), (3) ‘cọc’ (tiền cọc – deposit), (4) ‘cưỡng’ (cưỡng chế - force), (5) ‘container’ (xe container - container truck). For females: (1) ‘hò’ (hẹn hò, hát hò – date, sing) , (2) ‘dance’, (3) ‘piano’, (4) ‘lứt’ (gạo lứt - brown rice), (5) inox (đồ dùng inox - stainless steel utensils). None of these words are inherently gendered in Vietnamese grammar. Instead, they seem to reflect different topics or areas of interest that may be more common among males or females in the context of our dataset.

\clearpage
\onecolumn
\begin{longtable}{@{}c>{\raggedright\arraybackslash}p{2.4cm}ccccccc@{}}
\toprule
\footnotesize{\textbf{No.}} & \multicolumn{1}{c}{\footnotesize{\textbf{Province Name}}} & \footnotesize{\textbf{Region}} & \footnotesize{\textbf{Province Code}} & \footnotesize{\textbf{Duration}} & \footnotesize{\textbf{Speakers}} & \footnotesize{\textbf{Records}} & \footnotesize{\textbf{Words}} & \footnotesize{\textbf{Unique Words}} \\
\midrule
\endfirsthead

\multicolumn{9}{c}%
{{\tablename\ \thetable{} -- continued from previous page}} \\
\toprule
\footnotesize{\textbf{No.}} & \multicolumn{1}{c}{\footnotesize{\textbf{Province Name}}} & \footnotesize{\textbf{Region}} & \footnotesize{\textbf{Province Code}} & \footnotesize{\textbf{Duration}} & \footnotesize{\textbf{Speakers}} & \footnotesize{\textbf{Records}} & \footnotesize{\textbf{Words}} & \footnotesize{\textbf{Unique Words}} \\
\midrule
\endhead

\midrule \multicolumn{9}{r}{{Continued on next page}} \\
\endfoot
\bottomrule
\caption[]{List of 63 Provinces of Vietnam with Language Data Statistics.}
\label{tab: Statistic_province}
\endlastfoot

        \footnotesize{1} & \footnotesize{Cao Bằng} & \footnotesize{North} & \footnotesize{11} & \footnotesize{101.97} & \footnotesize{158} & \footnotesize{357} & \footnotesize{19,621} & \footnotesize{1,355} \\
        \footnotesize{2} & \footnotesize{Lạng Sơn} & \footnotesize{North} & \footnotesize{12} & \footnotesize{99.48} & \footnotesize{171} & \footnotesize{304} & \footnotesize{20,304} & \footnotesize{1,317} \\
        \footnotesize{3} & \footnotesize{Quảng Ninh} & \footnotesize{North} & \footnotesize{14} & \footnotesize{99.77} & \footnotesize{251} & \footnotesize{316} & \footnotesize{20,282} & \footnotesize{1,423} \\
        \footnotesize{4} & \footnotesize{Hải Phòng} & \footnotesize{North} & \footnotesize{15} & \footnotesize{98.72} & \footnotesize{223} & \footnotesize{306} & \footnotesize{20,714} & \footnotesize{1,545} \\
        \footnotesize{5} & \footnotesize{Thái Bình} & \footnotesize{North} & \footnotesize{17} & \footnotesize{100.96} & \footnotesize{300} & \footnotesize{329} & \footnotesize{21,488} & \footnotesize{1,534} \\
        \footnotesize{6} & \footnotesize{Nam Định} & \footnotesize{North} & \footnotesize{18} & \footnotesize{99.38} & \footnotesize{151} & \footnotesize{300} & \footnotesize{20,403} & \footnotesize{1,143} \\
        \footnotesize{7} & \footnotesize{Phú Thọ} & \footnotesize{North} & \footnotesize{19} & \footnotesize{106.99} & \footnotesize{273} & \footnotesize{313} & \footnotesize{22,407} & \footnotesize{1,483} \\
        \footnotesize{8} & \footnotesize{Thái Nguyên} & \footnotesize{North} & \footnotesize{20} & \footnotesize{97.45} & \footnotesize{230} & \footnotesize{314} & \footnotesize{20,780} & \footnotesize{1,438} \\
        \footnotesize{9} & \footnotesize{Yên Bái} & \footnotesize{North} & \footnotesize{21} & \footnotesize{98.98} & \footnotesize{196} & \footnotesize{289} & \footnotesize{20,187} & \footnotesize{1,376} \\
        \footnotesize{10} & \footnotesize{Tuyên Quang} & \footnotesize{North} & \footnotesize{22} & \footnotesize{98.59} & \footnotesize{199} & \footnotesize{286} & \footnotesize{20,595} & \footnotesize{1,124} \\
        \footnotesize{11} & \footnotesize{Hà Giang} & \footnotesize{North} & \footnotesize{23} & \footnotesize{100.56} & \footnotesize{309} & \footnotesize{325} & \footnotesize{20,401} & \footnotesize{1,330} \\
        \footnotesize{12} & \footnotesize{Lào Cai} & \footnotesize{North} & \footnotesize{24} & \footnotesize{98.38} & \footnotesize{286} & \footnotesize{313} & \footnotesize{19,754} & \footnotesize{1,377} \\
        \footnotesize{13} & \footnotesize{Lai Châu} & \footnotesize{North} & \footnotesize{25} & \footnotesize{99.46} & \footnotesize{172} & \footnotesize{295} & \footnotesize{20,007} & \footnotesize{1,233} \\
        \footnotesize{14} & \footnotesize{Sơn La} & \footnotesize{North} & \footnotesize{26} & \footnotesize{99.53} & \footnotesize{244} & \footnotesize{308} & \footnotesize{19,702} & \footnotesize{1,399} \\
        \footnotesize{15} & \footnotesize{Điện Biên} & \footnotesize{North} & \footnotesize{27} & \footnotesize{97.07} & \footnotesize{191} & \footnotesize{283} & \footnotesize{19,025} & \footnotesize{1,316} \\
        \footnotesize{16} & \footnotesize{Hòa Bình} & \footnotesize{North} & \footnotesize{28} & \footnotesize{97.89} & \footnotesize{187} & \footnotesize{281} & \footnotesize{19,406} & \footnotesize{1,322} \\
        \footnotesize{17} & \footnotesize{Hà Nội} & \footnotesize{North} & \footnotesize{30} & \footnotesize{94.82} & \footnotesize{190} & \footnotesize{316} & \footnotesize{19,930} & \footnotesize{1,531} \\
        \footnotesize{18} & \footnotesize{Hải Dương} & \footnotesize{North} & \footnotesize{34} & \footnotesize{96.59} & \footnotesize{196} & \footnotesize{290} & \footnotesize{19,730} & \footnotesize{1,451} \\
        \footnotesize{19} & \footnotesize{Ninh Bình} & \footnotesize{North} & \footnotesize{35} & \footnotesize{96.57} & \footnotesize{250} & \footnotesize{268} & \footnotesize{20,143} & \footnotesize{1,441} \\
        \footnotesize{20} & \footnotesize{Vĩnh Phúc} & \footnotesize{North} & \footnotesize{88} & \footnotesize{92.71} & \footnotesize{208} & \footnotesize{266} & \footnotesize{18,543} & \footnotesize{1,280} \\
        \footnotesize{21} & \footnotesize{Hưng Yên} & \footnotesize{North} & \footnotesize{89} & \footnotesize{89.64} & \footnotesize{211} & \footnotesize{277} & \footnotesize{17,940} & \footnotesize{1,473} \\
        \footnotesize{22} & \footnotesize{Hà Nam} & \footnotesize{North} & \footnotesize{90} & \footnotesize{90.48} & \footnotesize{213} & \footnotesize{272} & \footnotesize{18,229} & \footnotesize{1,290} \\
        \footnotesize{23} & \footnotesize{Bắc Kạn} & \footnotesize{North} & \footnotesize{97} & \footnotesize{97.02} & \footnotesize{208} & \footnotesize{291} & \footnotesize{19,169} & \footnotesize{1,341} \\
        \footnotesize{24} & \footnotesize{Bắc Giang} & \footnotesize{North} & \footnotesize{98} & \footnotesize{91.44} & \footnotesize{207} & \footnotesize{272} & \footnotesize{17,998} & \footnotesize{1,315} \\
        \footnotesize{25} & \footnotesize{Bắc Ninh} & \footnotesize{North} & \footnotesize{99} & \footnotesize{93.53} & \footnotesize{193} & \footnotesize{277} & \footnotesize{18,956} & \footnotesize{1,355} \\
        \footnotesize{26} & \footnotesize{Thanh Hóa} & \footnotesize{Central} & \footnotesize{36} & \footnotesize{97.19} & \footnotesize{188} & \footnotesize{303} & \footnotesize{20,068} & \footnotesize{1,399} \\
        \footnotesize{27} & \footnotesize{Nghệ An} & \footnotesize{Central} & \footnotesize{37} & \footnotesize{117.98} & \footnotesize{275} & \footnotesize{363} & \footnotesize{24,549} & \footnotesize{1,410} \\
        \footnotesize{28} & \footnotesize{Hà Tĩnh} & \footnotesize{Central} & \footnotesize{38} & \footnotesize{100.35} & \footnotesize{226} & \footnotesize{305} & \footnotesize{20,472} & \footnotesize{1,487} \\
        \footnotesize{29} & \footnotesize{Đà Nẵng} & \footnotesize{Central} & \footnotesize{43} & \footnotesize{101.82} & \footnotesize{253} & \footnotesize{337} & \footnotesize{21,460} & \footnotesize{1,430} \\
        \footnotesize{30} & \footnotesize{Đắk Lắk} & \footnotesize{Central} & \footnotesize{47} & \footnotesize{98.54} & \footnotesize{220} & \footnotesize{306} & \footnotesize{18,935} & \footnotesize{1,456} \\
        \footnotesize{31} & \footnotesize{Đắk Nông} & \footnotesize{Central} & \footnotesize{48} & \footnotesize{99.79} & \footnotesize{263} & \footnotesize{304} & \footnotesize{20,296} & \footnotesize{1,498} \\
        \footnotesize{32} & \footnotesize{Lâm Đồng} & \footnotesize{Central} & \footnotesize{49} & \footnotesize{99.05} & \footnotesize{179} & \footnotesize{303} & \footnotesize{19,416} & \footnotesize{1,408} \\
        \footnotesize{33} & \footnotesize{Quảng Bình} & \footnotesize{Central} & \footnotesize{73} & \footnotesize{100.94} & \footnotesize{208} & \footnotesize{345} & \footnotesize{20,857} & \footnotesize{1,638} \\
        \footnotesize{34} & \footnotesize{Quảng Trị} & \footnotesize{Central} & \footnotesize{74} & \footnotesize{91.19} & \footnotesize{143} & \footnotesize{292} & \footnotesize{17,893} & \footnotesize{1,365} \\
        \footnotesize{35} & \footnotesize{Thừa Thiên Huế} & \footnotesize{Central} & \footnotesize{75} & \footnotesize{101.04} & \footnotesize{187} & \footnotesize{303} & \footnotesize{1,9870} & \footnotesize{1,403} \\
        \footnotesize{36} & \footnotesize{Quảng Ngãi} & \footnotesize{Central} & \footnotesize{76} & \footnotesize{98.29} & \footnotesize{242} & \footnotesize{330} & \footnotesize{20,432} & \footnotesize{1,627} \\
        \footnotesize{37} & \footnotesize{Bình Định} & \footnotesize{Central} & \footnotesize{77} & \footnotesize{98.13} & \footnotesize{169} & \footnotesize{307} & \footnotesize{20,754} & \footnotesize{1,428} \\
        \footnotesize{38} & \footnotesize{Phú Yên} & \footnotesize{Central} & \footnotesize{78} & \footnotesize{99.54} & \footnotesize{249} & \footnotesize{305} & \footnotesize{19,841} & \footnotesize{1,525} \\
        \footnotesize{39} & \footnotesize{Khánh Hòa} & \footnotesize{Central} & \footnotesize{79} & \footnotesize{97} & \footnotesize{182} & \footnotesize{281} & \footnotesize{19,522} & \footnotesize{1,352} \\
        \footnotesize{40} & \footnotesize{Gia Lai} & \footnotesize{Central} & \footnotesize{81} & \footnotesize{99.66} & \footnotesize{267} & \footnotesize{314} & \footnotesize{19,696} & \footnotesize{1,361} \\
        \footnotesize{41} & \footnotesize{Kon Tum} & \footnotesize{Central} & \footnotesize{82} & \footnotesize{96.52} & \footnotesize{192} & \footnotesize{305} & \footnotesize{17,670} & \footnotesize{1,439} \\
        \footnotesize{42} & \footnotesize{Ninh Thuận} & \footnotesize{Central} & \footnotesize{85} & \footnotesize{97.94} & \footnotesize{235} & \footnotesize{314} & \footnotesize{19,789} & \footnotesize{1,426} \\
        \footnotesize{43} & \footnotesize{Bình Thuận} & \footnotesize{Central} & \footnotesize{86} & \footnotesize{101.58} & \footnotesize{211} & \footnotesize{325} & \footnotesize{20,455} & \footnotesize{1,575} \\
        \footnotesize{44} & \footnotesize{Quảng Nam} & \footnotesize{Central} & \footnotesize{92} & \footnotesize{94.94} & \footnotesize{190} & \footnotesize{291} & \footnotesize{19,526} & \footnotesize{1,527} \\
        \footnotesize{45} & \footnotesize{Hồ Chí Minh} & \footnotesize{South} & \footnotesize{59} & \footnotesize{91.68} & \footnotesize{276} & \footnotesize{318} & \footnotesize{19,823} & \footnotesize{1,349} \\
        \footnotesize{46} & \footnotesize{Đồng Nai} & \footnotesize{South} & \footnotesize{60} & \footnotesize{92.78} & \footnotesize{142} & \footnotesize{275} & \footnotesize{18,738} & \footnotesize{1,422} \\
        \footnotesize{47} & \footnotesize{Bình Dương} & \footnotesize{South} & \footnotesize{61} & \footnotesize{99.66} & \footnotesize{242} & \footnotesize{314} & \footnotesize{20,431} & \footnotesize{1,407} \\
        \footnotesize{48} & \footnotesize{Long An} & \footnotesize{South} & \footnotesize{62} & \footnotesize{97.68} & \footnotesize{241} & \footnotesize{308} & \footnotesize{19,576} & \footnotesize{1,505} \\
        \footnotesize{49} & \footnotesize{Tiền Giang} & \footnotesize{South} & \footnotesize{63} & \footnotesize{94.57} & \footnotesize{154} & \footnotesize{289} & \footnotesize{18,629} & \footnotesize{1,467} \\
        \footnotesize{50} & \footnotesize{Vĩnh Long} & \footnotesize{South} & \footnotesize{64} & \footnotesize{98.67} & \footnotesize{142} & \footnotesize{284} & \footnotesize{19,820} & \footnotesize{1,397} \\
        \footnotesize{51} & \footnotesize{Cần Thơ} & \footnotesize{South} & \footnotesize{65} & \footnotesize{89.11} & \footnotesize{213} & \footnotesize{263} & \footnotesize{16,970} & \footnotesize{1,175} \\
        \footnotesize{52} & \footnotesize{Đồng Tháp} & \footnotesize{South} & \footnotesize{66} & \footnotesize{93.64} & \footnotesize{196} & \footnotesize{273} & \footnotesize{19,281} & \footnotesize{1,409} \\
        \footnotesize{53} & \footnotesize{An Giang} & \footnotesize{South} & \footnotesize{67} & \footnotesize{97.65} & \footnotesize{135} & \footnotesize{285} & \footnotesize{18,929} & \footnotesize{1,409} \\
        \footnotesize{54} & \footnotesize{Kiên Giang} & \footnotesize{South} & \footnotesize{68} & \footnotesize{97.21} & \footnotesize{117} & \footnotesize{278} & \footnotesize{18,521} & \footnotesize{1,505} \\
        \footnotesize{55} & \footnotesize{Cà Mau} & \footnotesize{South} & \footnotesize{69} & \footnotesize{99.74} & \footnotesize{213} & \footnotesize{302} & \footnotesize{19,097} & \footnotesize{1,500} \\
        \footnotesize{56} & \footnotesize{Tây Ninh} & \footnotesize{South} & \footnotesize{70} & \footnotesize{100.15} & \footnotesize{160} & \footnotesize{302} & \footnotesize{20,052} & \footnotesize{1,316} \\
        \footnotesize{57} & \footnotesize{Bến Tre} & \footnotesize{South} & \footnotesize{71} & \footnotesize{97.98} & \footnotesize{143} & \footnotesize{289} & \footnotesize{18,627} & \footnotesize{1,378} \\
        \footnotesize{58} & \footnotesize{Bà Rịa - Vũng Tàu} & \footnotesize{South} & \footnotesize{72} & \footnotesize{100.45} & \footnotesize{277} & \footnotesize{319} & \footnotesize{19,935} & \footnotesize{1,461} \\
        \footnotesize{59} & \footnotesize{Sóc Trăng} & \footnotesize{South} & \footnotesize{83} & \footnotesize{92.95} & \footnotesize{131} & \footnotesize{273} & \footnotesize{17,389} & \footnotesize{1,326} \\
        \footnotesize{60} & \footnotesize{Trà Vinh} & \footnotesize{South} & \footnotesize{84} & \footnotesize{98.39} & \footnotesize{138} & \footnotesize{293} & \footnotesize{19,050} & \footnotesize{1,383} \\
        \footnotesize{61} & \footnotesize{Bình Phước} & \footnotesize{South} & \footnotesize{93} & \footnotesize{98.57} & \footnotesize{215} & \footnotesize{319} & \footnotesize{19,319} & \footnotesize{1,559} \\
        \footnotesize{62} & \footnotesize{Bạc Liêu} & \footnotesize{South} & \footnotesize{94} & \footnotesize{93.99} & \footnotesize{88} & \footnotesize{278} & \footnotesize{17,851} & \footnotesize{1,257} \\
        \footnotesize{63} & \footnotesize{Hậu Giang} & \footnotesize{South} & \footnotesize{95} & \footnotesize{95.38} & \footnotesize{246} & \footnotesize{309} & \footnotesize{19,756} & \footnotesize{1,419} \\

\end{longtable}
\twocolumn
\clearpage
\section{Experimental Settings}
\label{sec: Appendix_C}

The pretrained models were originally trained for the SR task. Therefore, when fine-tuning them for the DI task, we add two linear layers on top of the pretrained models and the cross-entropy loss function during the training.

The models tasked with Dialect Identification are configured with the hyperparameters listed in \autoref{tab: DI_Cofig}, while the models responsible for Speech Recognition utilize the hyperparameters detailed in \autoref{tab: SR_Cofig}. All experimental training are carried out on an NVIDIA GeForce RTX 4090 (24GB).

\begin{table}[h!]
    \centering
    \resizebox{\columnwidth}{!}{%
    \begin{tabular}{lcc|c|c}
        \toprule
        \multirow{2}{*}{\textbf{Hyperparameter}} & \multicolumn{2}{c|}{\textbf{wav2vec 2.0}} & \multirow{2}{*}{\textbf{XLSR}} & \multirow{2}{*}{\textbf{Whisper}} \\
        \cmidrule{2-3}
        & \textbf{Base} & \textbf{Large} & \textbf{XLS-R} \\
        \midrule
        Epochs & 15 & 15 & 15 & 15 \\
        Learning rate & 3e-5 & 6e-5 & 6e-5 & 3e-5 \\
        Batch size & 64 & 64 & 64 & 64\\
        Optimizer & AdamW & AdamW & AdamW & AdamW \\
        Weight Decay & 0 & 0 & 0 & 0 \\
        Warmup Ratio & 0.1 & 0.1 & 0.1 & 0.1 \\
        \bottomrule
    \end{tabular}
    }
    \caption{Dialect Identification experimental configurations.}
    \label{tab: DI_Cofig}
\end{table}

\begin{table}[h!]
    \centering    
    \begin{tabular}{lc|c}
        \toprule
        \textbf{Hyperparameter} & \textbf{wav2vec 2.0} &\textbf{Whisper} \\
        \midrule
        Epochs & 15 & 10 \\
        Learning rate & 1e-4 & 1e-5 \\
        Batch size & 8 & 8 \\
        Optimizer & AdamW & AdamW \\
        Weight Decay & 0.005 & 0.005 \\
        Warmup Ratio & 0.1 & 0.1 \\
        \bottomrule
    \end{tabular}
    \caption{Speech Recognition experimental configurations.}
    \label{tab: SR_Cofig}
\end{table}

\section{Evaluation Metrics}
\label{sec: Detail_Metrics}
\subsection{F1-macro}

The F1-score \cite{f1_macro} is a metric used in statistical machine learning to evaluate the performance of a classification model. The F1-score is calculated based on precision and recall, and is the harmonic mean of these two measures. The formula for the F1 score is given by:

\[
F_1 = \frac{2 \times \text{precision} \times \text{recall}}{\text{precision} + \text{recall}}
\]

To calculate the macro F1-score, the F1 score is computed for each class individually, and then the average of these F1-scores is taken. The formula for the macro F1-score is as follows:

\[
F_{1 \text{ macro}} = \frac{1}{n} \sum_{i=1}^{n} F_{1i}
\]

where \( n \) is the number of classes and \( F_{1i} \) is the F1 score for the \( i \)-th class. The macro F1-score is not affected by imbalances in class distribution, as each class is treated equally when averaging.

In our dialect identification tasks, we choose to use the macro F1-score as the evaluation metric to ensure that the performance of the classification model across each provincial dialect is computed fairly, without being influenced by the disparity in sample sizes among the provinces. This is particularly important in this study, where each province represents a distinct dialect that needs to be treated with equivalent fairness.

\subsection{WER}

Word Error Rate (WER) \cite{WER} is a crucial metric used to evaluate the performance of Speech Recognition systems. It is measured based on the accuracy of the transcription generated by the system compared to the reference transcription, by considering three types of errors: substitution errors (S), deletion errors (D), and insertion errors (I), relative to the total number of words in the reference transcript (N). The formula for the WWER is as follows:

\[
WER = \frac{S + D + I}{N}
\]

\section{Experimental Results}
\label{sec: Appendix_D}
\subsection{Dialect Identification}

\begin{figure}[h]
    \centering
    \includegraphics[width=0.48\textwidth]{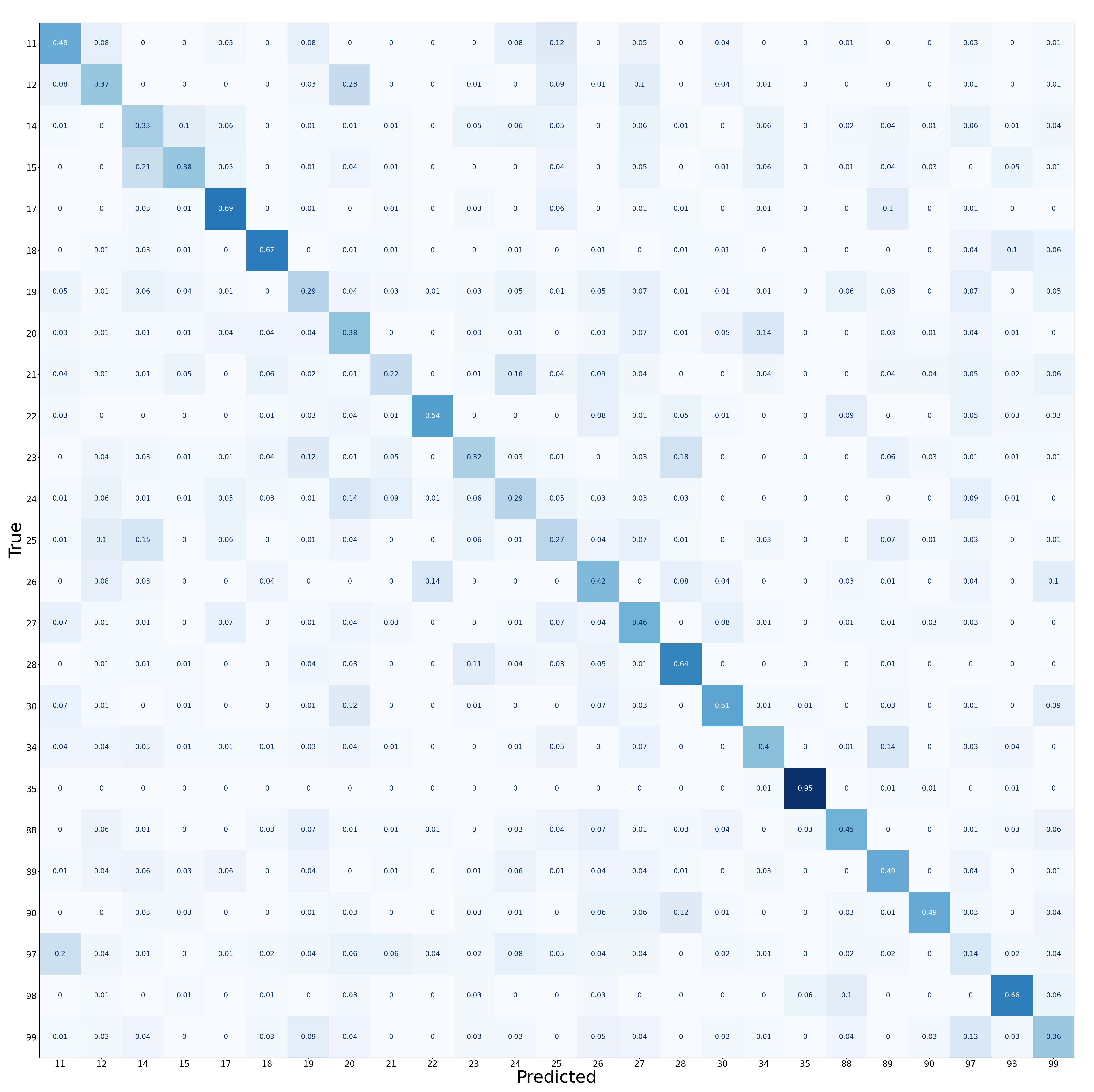}
    \caption{Confusion matrix of [DI\_North].}
    \label{fig:result_DI_North}
\end{figure}

\begin{figure}[h]
    \centering
    \includegraphics[width=0.46\textwidth]{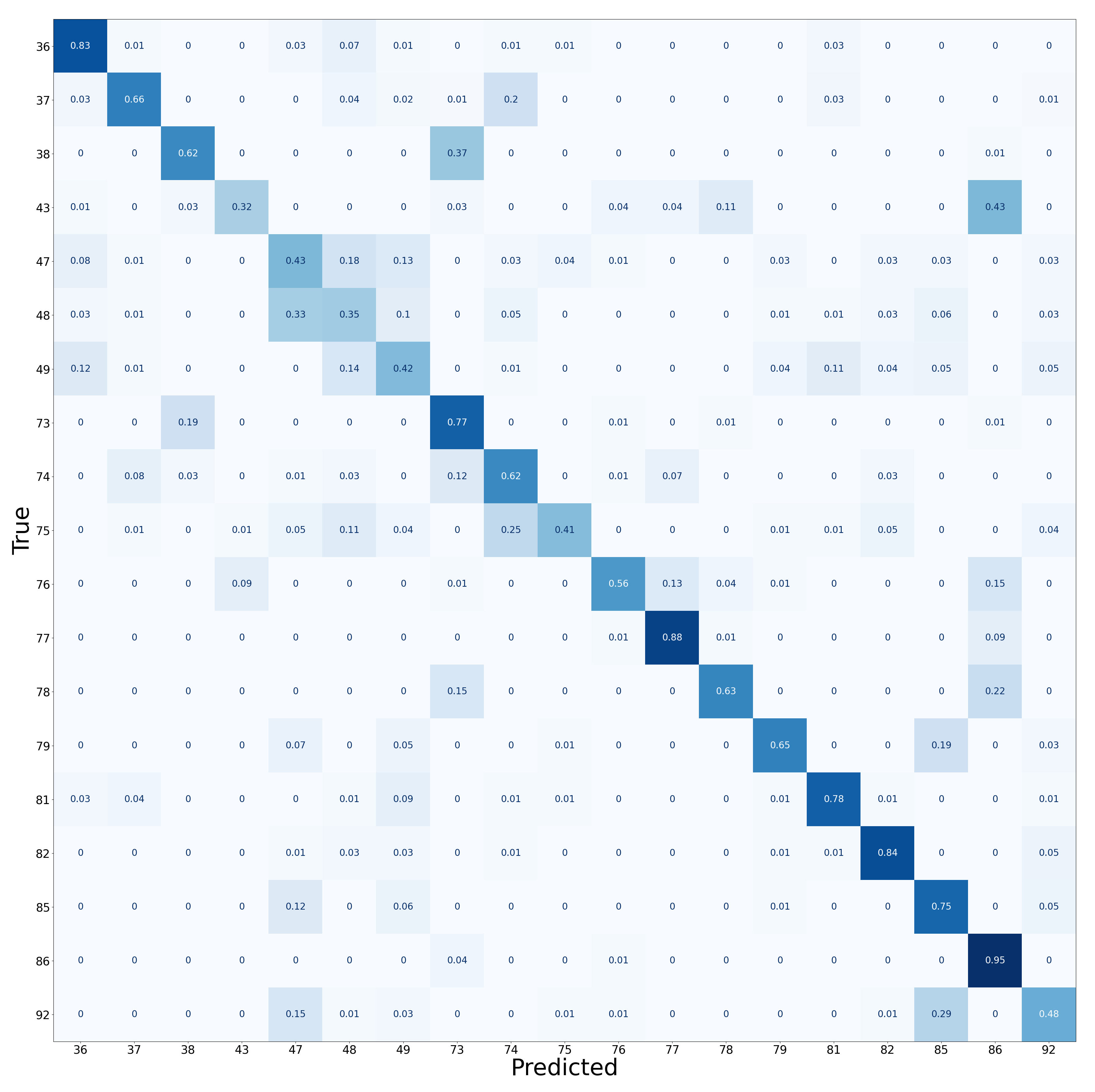}
    \caption{Confusion matrix of [DI\_Central].}
    \label{fig:result_DI_Central}
\end{figure}

\begin{figure}[h]
    \centering
    \includegraphics[width=0.48\textwidth]{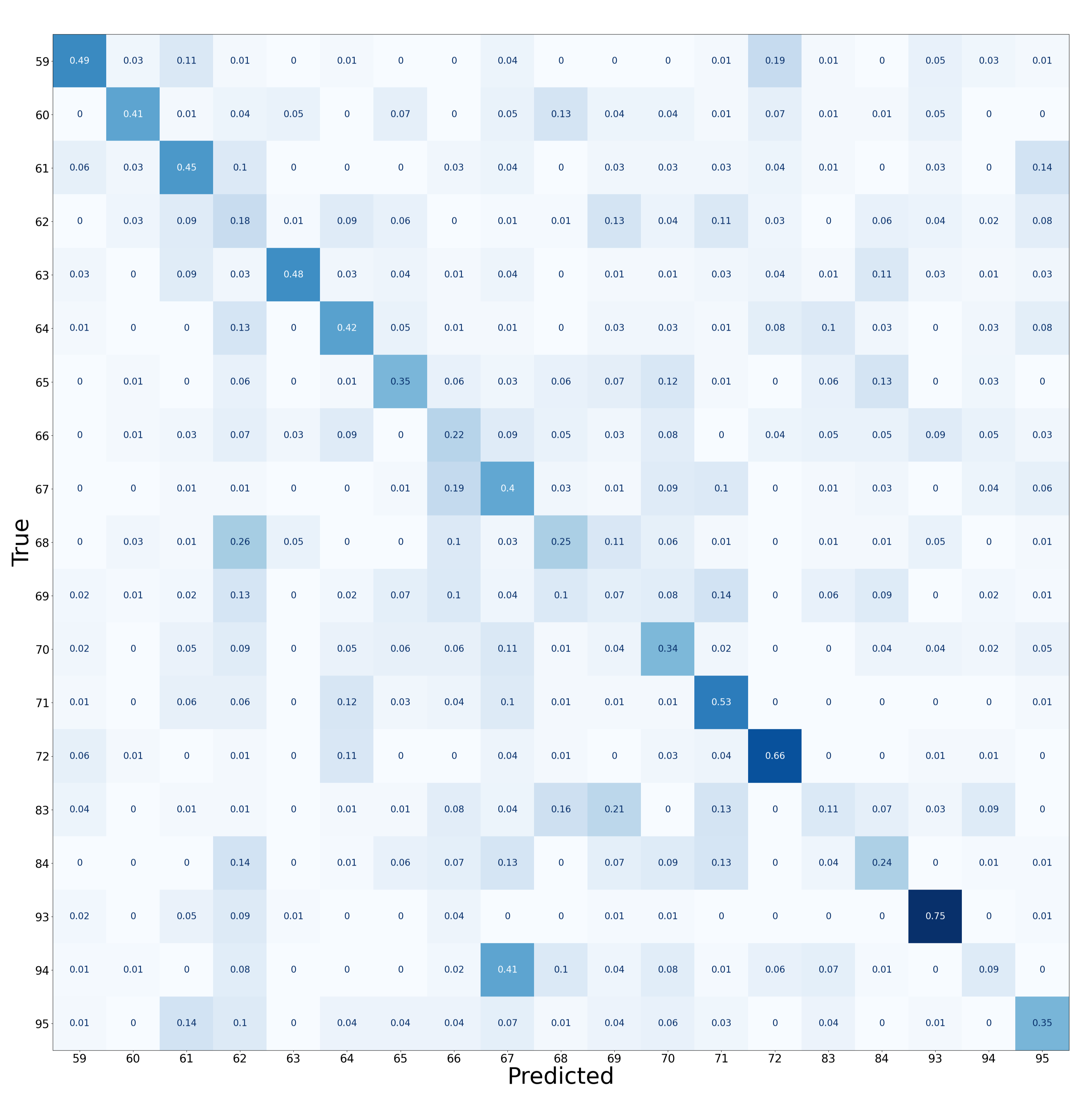}
    \caption{Confusion matrix of [DI\_South].}
    \label{fig:result_DI_South}
\end{figure}

\begin{figure}[h]
    \centering
    \includegraphics[width=0.48\textwidth]{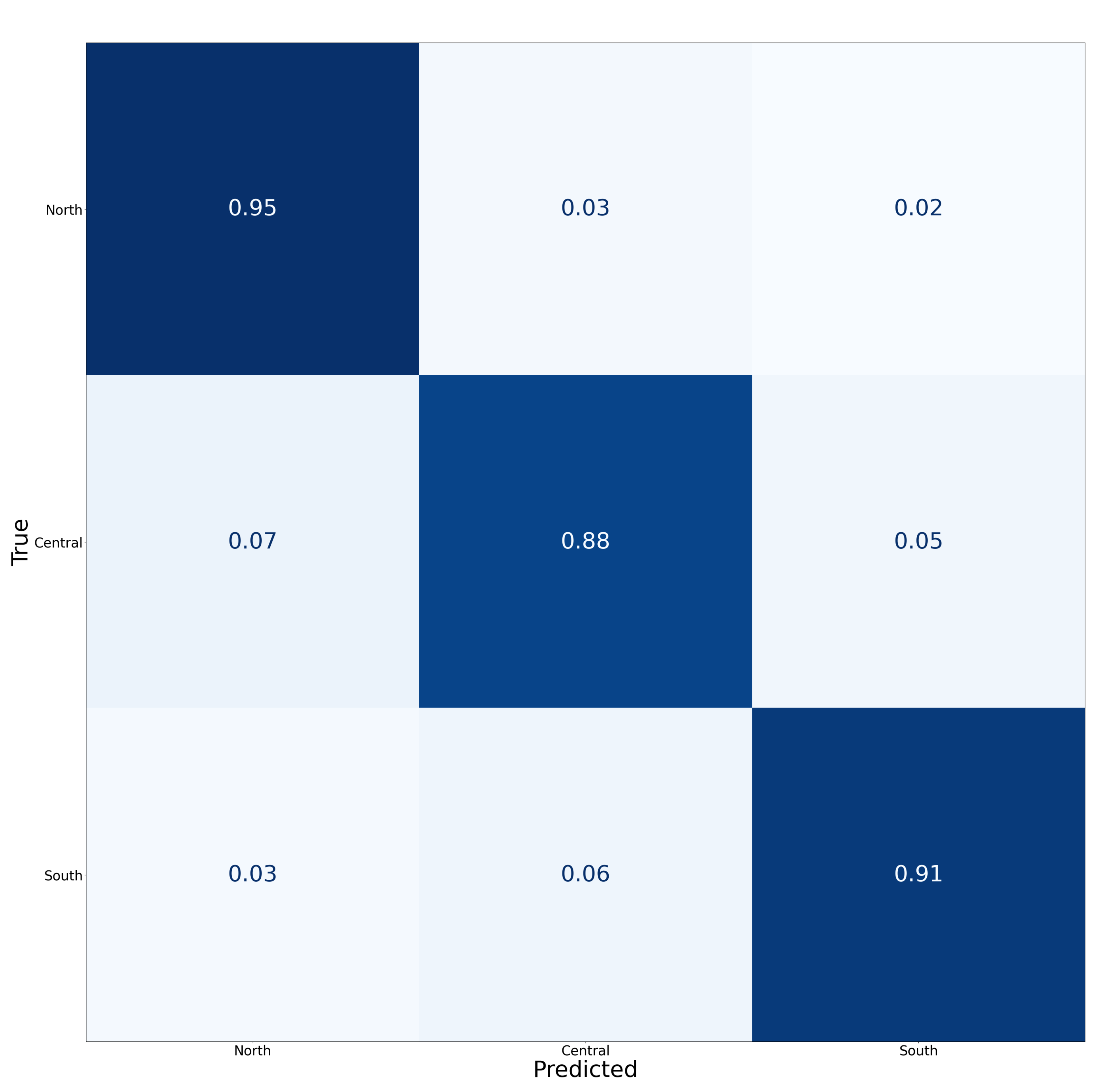}
    \caption{Confusion matrix of [DI\_VN\_3].}
    \label{fig:result_DI_VN3}
\end{figure}

\begin{figure*}[h]
    \centering
    \includegraphics[width=\textwidth]{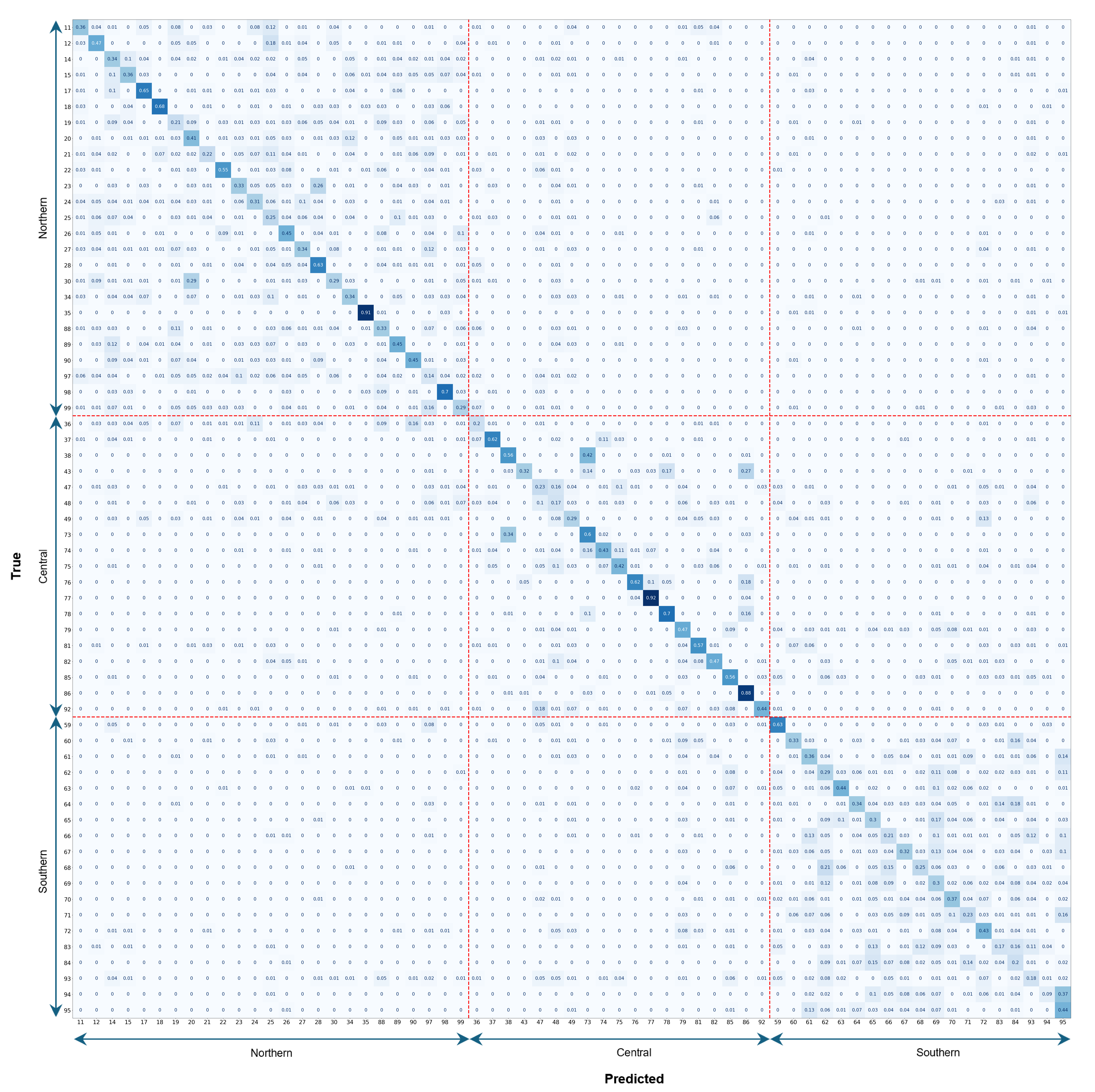}
    \caption{Confusion matrix of [DI\_VN\_63].}
    \label{fig: Appendix_4_result_DI_VN63}
\end{figure*}

The presented results originate from the models exhibiting the highest performance for each respective task, as illustrated in \autoref{fig:result_DI_North}, \autoref{fig:result_DI_Central}, \autoref{fig:result_DI_South}, \autoref{fig:result_DI_VN3} and \autoref{fig: Appendix_4_result_DI_VN63}. All confusion matrices are normalized with respect to the true conditions.

The results of the \textbf{[DI\_VN\_3]} experiment demonstrate the model's remarkable dialect identification performance across all three regions, correctly identifying 95\%, 88\%, and 91\% of the samples for the Northern, Central, and Southern dialects, respectively. However, in the \textbf{[DI\_North]} task, the accuracy rates for predicting provincial dialects were uneven, ranging from a high of 95\% (Ninh Binh - 35) for the top-performing province to a low of 14\% (Bac Giang - 14) for the least accurate one. The largest confusion occurred for label 12 (Lang Son), which was predicted as label 20 (Thai Nguyen) with an error rate of up to 23\%. The \textbf{[DI\_Central]} task demonstrates relatively promising recognition rates for many provinces, with 12 out of the 19 provinces achieving accurate predictions for over 60\% of their samples. However, notable confusion persists among certain geographically close provinces. As an example, the province of Dak Nong (48) is frequently misclassified as Dak Lak (47), and Ha Tinh (38) is often predicted as Quang Binh (73). Against all expectations, Da Nang City (43) is predicted as Binh Thuan (86) with the highest confusion rate of 43\%, despite the substantial geographical distance separating the two provinces, making this finding quite inexplicable. In the \textbf{[DI\_South]} dialect identification task, substantial confusion was observed among provincial dialects, potentially attributable to the high degree of similarity between them. In particular, the provincial dialects of Ca Mau (69) and Bac Lieu (94) have very low correct identification rates of only 7\% and 9\%, respectively. The confusion matrix also indicates that the two provincial dialects of Binh Phuoc (93) and Ba Ria - Vung Tau (72) are the most accurately classified, with rates of 75\% and 66\%, respectively.

The outcomes of the \textbf{[DI\_VN\_63]} experiment are depicted in \autoref{fig: Appendix_4_result_DI_VN63}. We have incorporated red dashed lines to facilitate the tracking of provincial dialects across regional dialect boundaries. Overall, the confusion between provincial dialects is primarily concentrated within the three regional dialect clusters. While some confusion persists between geographically proximate regional dialects, such as Northern - Central and Central - Southern, the most geographically distant pair, Northern - Southern, exhibits the least confusion. Among Vietnam's five municipalities of Vietnam, four cities – Ha Noi (30), Hai Phong (15), Da Nang (43), and Can Tho (65) - have very low prediction accuracy rates, ranging from 29\% to 36\%, which can be explained by the influx of residents from other provinces; however, surprisingly, Ho Chi Minh (59) has a very high prediction accuracy rate of 63\%. The neighboring provinces of Ha Tinh (38) and Quang Binh (73) exhibit a high rate of mutual misprediction, with 42\% of samples of label 38 being predicted as 73, and in 34\% of cases, label 73 is predicted as 38. The three most distinctive provinces are Binh Dinh (77), Ninh Binh (35), and Binh Thuan (86), with prediction accuracy rates of 92\%, 91\%, and 88\%, respectively.

\subsection{Model Improvement with Training on Entire Data}

We analyze the results from four dialectal ASR experiments. For each experiment, we select the best-performing model based on WER. The wav2vec2-base-vi-vlsp2020 model performs best in the [SR\_North] and [SR\_VN\_63] experiments, the wav2vec2-base-vietnamese-250h model outperforms others in the [SR\_Central] experiment, and the PhoWhisperbase model achieves the top performance in the [SR\_South] experiment. The error analysis focuses on the improvement in errors when trained on a combined dataset containing all dialects.

\begin{figure*}[h]
    \centering
    \includegraphics[width=1.0\linewidth]{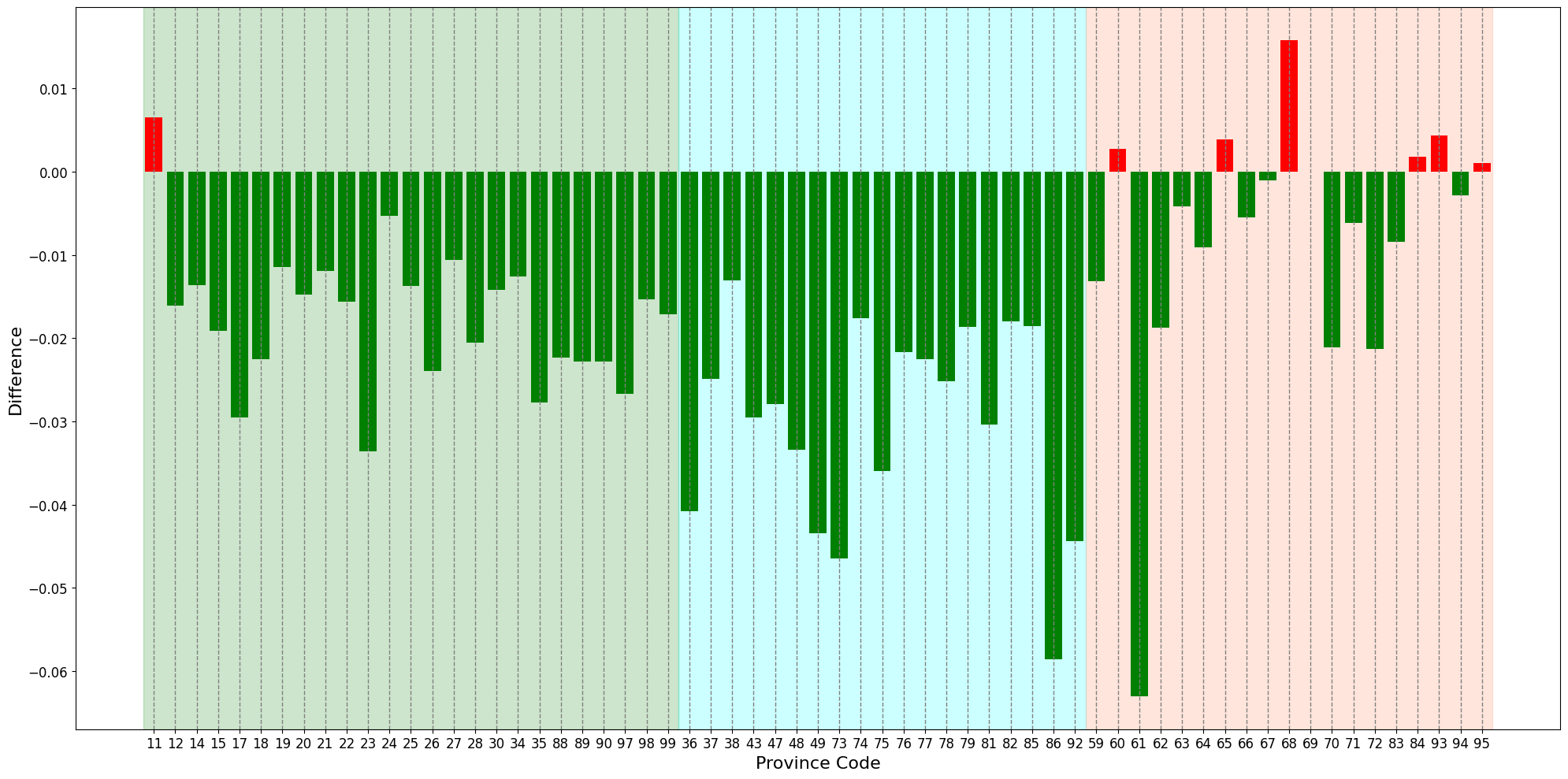}
    \caption{WER discrepancy when fine-tuning the model on the entire dataset versus three sub-datasets.}
    \label{fig: Appendix_Compare_WER_best_VN63}
\end{figure*}

The examples are presented in \autoref{tab: Appendix_SR_Error_Improvement}. The red text in the `Regional Data' column indicates errors when trained on a specific regional dialect. For the `Entire Data' column, red text represents errors that were not resolved, orange text indicates errors that were partially resolved but not entirely, and green text denotes errors that were completely resolved. The models trained on the entire Vietnamese dataset perform better than those trained only on specific regional dialects. For instance, the confusion between `d' and `gi' in the Northern dialect was resolved. In the Central Dialect, although the spelling was not perfectly accurate, the model's prediction more closely mirrored the original phonetics than the model from the [SR\_Central] experiment. However, this improvement was still quite limited, as exemplified by the sample from the Southern dialect. \autoref{fig: Appendix_Compare_WER_best_VN63} presents the Word Error Rate (WER) discrepancy when the model is fine-tuned on the entire dataset and fine-tuned on three sub-datasets. Red indicates instances where fine-tuning on the entire dataset performs worse, while blue indicates instances where fine-tuning on the entire dataset performs better.

\begin{table*}[htbp]
\centering
\small
\begin{tabular}{p{0.6in}|p{0.6in}|p{0.95in}|p{0.95in}|p{0.951in}|p{0.95in}}
\toprule
\multicolumn{1}{c|}{\textbf{Task}} & \multicolumn{1}{c|}{\textbf{Province code}} & \multicolumn{1}{c|}{\textbf{Reference Transcript}} & \multicolumn{1}{c|}{\textbf{Without Fine-tuned}} & \multicolumn{1}{c|}{\textbf{Fine-tuned}} & \multicolumn{1}{c}{\textbf{English}} \\

\midrule

\multirow{2}{*}{[SR\_North]} & 24 & năm ngày nay là tất cả các nhà máy đang ngừng hoạt động ở lào cai rồi ninh bình thanh hóa hưng yên ngừng hết & năm ngày nay là tất cả các nhà máy đang ngừng hoạt động ở lào cai rồi \textcolor{red}{linh bình} thanh hóa hưng yên ngừng hết & năm ngày nay là tất cả các nhà máy đang ngừng hoạt động ở lào cai rồi \textcolor{green}{ninh bình} thanh hóa hưng yên ngừng hết & For the past five days. all factories in Lào Cai, Ninh Bình, Thanh Hóa, and Hưng Yên have been shut down. \\
\\
& 25 & người lao động thôn được đào tạo nghề đã tìm được việc làm mới & người lao động thôn được đào tạo nghề thì đã tìm được \textcolor{red}{việt nàm} mới & người lao động thôn được đào tạo nghề đã tìm được \textcolor{green}{việc làm} mới & The villager workers who were trained in vocational skills have found new jobs. \\

\midrule

\multirow{3}{*}{[SR\_Central]} & 75 & các đối tượng là thương binh bệnh binh & \textcolor{red}{két} đối tượng là thương binh bệnh binh & \textcolor{green}{các} đối tượng là thương binh bệnh binh & The individuals are war invalids and sick soldiers. \\
\\
& 76 & tất cả là trâu bò phải bán để cho cháu đi chữa bệnh & tất cả là trâu bò phải \textcolor{red}{bón} để cho cháu đi chữa bệnh & tất cả là trâu bò phải \textcolor{green}{bán} để cho cháu đi chữa bệnh & All the buffaloes and cows had to be sold to pay for the child's medical treatment. \\
\\
& 77 & đã hai năm rồi nhưng mà cũng không có thấy công ty & đã hai \textcolor{red}{nem} rồi nhưng mà cũng không có thấy công ty & đã hai \textcolor{green}{năm} rồi nhưng mà cũng không có thấy công ty & It has been two years, but there is still no sign of the company. \\

\midrule

\multirow{2}{*}{[SR\_South]} & 67 & các anh công an làm sai định danh & các anh công an làm \textcolor{red}{xai} định danh & các anh công an làm \textcolor{green}{sai} định danh & The police officers made a mistake in the identification. \\
\\
& 94 & chúng tôi là có hai mươi mô hình nổi trội & chúng tôi là có hai mươi mô hình \textcolor{red}{nổi chội} & chúng tôi là có hai mươi mô hình \textcolor{green}{nổi trội} & We have twenty outstanding models.\\

\bottomrule

\end{tabular}
\caption{Errors of the best-performing models in [SR\_North], [SR\_Central], and [SR\_South] Experiments.}
\label{tab: Appendix_SR_Error_Vocab}
\end{table*}

\begin{table*}[htbp]
\centering
\small
\begin{tabular}{
    >{\raggedright\arraybackslash}m{1.15in}|
    >{\centering\arraybackslash}m{0.6in}|
    >{\raggedright\arraybackslash}m{1.15in}|
    >{\raggedright\arraybackslash}m{1.15in}|
    >{\raggedright\arraybackslash}m{1.15in}}
\toprule
\multicolumn{1}{c|}{\textbf{Reference Transcript}} & \multicolumn{2}{c|}{\textbf{Regional Data}} & \multicolumn{1}{c|}{\textbf{Entire Data}} & \multicolumn{1}{c}{\textbf{English}} \\
\cmidrule(lr){2-3} \cmidrule(lr){4-4}
\multicolumn{1}{c|}{} & \textbf{Experiment} & \multicolumn{1}{c|}{\textbf{Transcript}} & \multicolumn{1}{c|}{\textbf{Transcript}} &  \\
\midrule

độ \textbf{\textcolor{blue}{dày}} của lá là \textbf{\textcolor{blue}{dày}} hơn & [SR\_North] & độ \textbf{\textcolor{red}{giày}} của lá là \textbf{\textcolor{red}{giày}} hơn & độ \textbf{\textcolor{green}{dày}} của lá là \textbf{\textcolor{green}{dày}} hơn & The thickness of the leaf is thicker \\
\\
nghề này là \textbf{\textcolor{blue}{lúc rảnh rỗi}} là mình làm & [SR\_Central] & nghề này là \textbf{\textcolor{red}{rút rẻn rổi}} là mình làm & nghề này là \textbf{\textcolor{orange}{lút rảng}} \textbf{\textcolor{green}{rỗi}} là mình làm & During free time, I do this job \\
\\
chế biến xong rồi sẽ chia \textbf{\textcolor{blue}{lên khay}} & [SR\_South] & chế biến xong rồi sẽ chia \textbf{\textcolor{red}{lơn khai}} & chế biến xong rồi sẽ chia \textbf{\textcolor{green}{lên}} \textbf{\textcolor{red}{khai}} & After cooking, I'll put it into serving trays \\

\bottomrule

\end{tabular}
\caption{Speech recognition performance improvement of experiment [SR\_VN\_63] over remaining experiments.}
\label{tab: Appendix_SR_Error_Improvement}
\end{table*}

\end{document}